\documentclass{article}

\usepackage{arxiv}

\usepackage{algorithm}
\usepackage[noend]{algpseudocode}
\usepackage{amssymb}
\usepackage{amsmath}
\usepackage{amsfonts}
\usepackage{booktabs}
\usepackage{graphicx}
\usepackage{multicol, multirow}
\usepackage{array}
\usepackage{natbib}
\usepackage{url,hyperref,lineno,microtype,subcaption}
\usepackage{fancyhdr}

\DeclareMathOperator*{\argmin}{arg\,min}
\DeclareMathOperator*{\argmax}{arg\,max}

\makeatletter

\newcolumntype{P}[1]{>{\centering\arraybackslash}p{#1}}

\title{Avoiding Catastrophe: Active Dendrites Enable Multi-Task Learning in Dynamic Environments}

\author{ 
    Abhiram Iyer\thanks{Equal contribution.} \\
    Numenta \\
	Carnegie Mellon University \\
	\texttt{abhirami@andrew.cmu.edu} \\
	\And
	Karan Grewal$^*$ \\
	Numenta \\
	\texttt{kgrewal@numenta.com} \\
	\And
	Akash Velu$^*$ \\
	Stanford University \\
	\texttt{avelu@stanford.edu} \\
	\And
	Lucas Oliveira Souza \\
	Numenta \\
	\texttt{lsouza@numenta.com} \\
	\And
	Jeremy Forest \\
	Cornell University \\
	\texttt{jerem.forest@gmail.com} \\
	\And
	Subutai Ahmad$^*$\thanks{To whom correspondence should be addressed.} \\
	Numenta \\
	\texttt{sahmad@numenta.com} \\
}

\begin{document}

\renewcommand{\shorttitle}{Avoiding Catastrophe}

\maketitle

\newcommand{\context}{\boldsymbol{c}}
\newcommand{\p}{\boldsymbol{p}}
\renewcommand{\u}{\boldsymbol{u}}
\newcommand{\w}{\boldsymbol{w}}
\newcommand{\x}{\boldsymbol{x}}
\newcommand{\y}{\boldsymbol{y}}
\newcommand{\acc}[1]{#1\%}
\newcommand{\bacc}[1]{\textbf{#1\%}}

\begin{abstract}
\noindent A key challenge for AI is to build embodied systems that operate in dynamically changing environments. Such systems must adapt to changing task contexts and learn continuously. Although standard deep learning systems achieve state of the art results on static benchmarks, they often struggle in dynamic scenarios. In these settings, error signals from multiple contexts can interfere with one another, ultimately leading to a phenomenon known as catastrophic forgetting. In this article we investigate biologically inspired architectures as solutions to these problems. Specifically, we show that the biophysical properties of dendrites and local inhibitory systems enable networks to dynamically restrict and route information in a context-specific manner.
Our key contributions are as follows. First, we propose a novel artificial neural network architecture that incorporates active dendrites and sparse representations into the standard deep learning framework. Next, we study the performance of this architecture on two separate benchmarks requiring task-based adaptation: Meta-World, a multi-task reinforcement learning environment where a robotic agent must learn to solve a variety of manipulation tasks simultaneously; and a continual learning benchmark in which the model’s prediction task changes throughout training.
Analysis on both benchmarks demonstrates the emergence of overlapping but distinct and sparse subnetworks, allowing the system to fluidly learn multiple tasks with minimal forgetting.
Our neural implementation marks the first time a single architecture has achieved competitive results in both multi-task and continual learning settings.
Our research sheds light on how biological properties of neurons can inform deep learning systems to address dynamic scenarios that are typically impossible for traditional ANNs to solve.

\end{abstract}


\section{Introduction}

Creating embodied systems that thrive in dynamically changing environments is a fundamental challenge for building intelligent systems. Humans handle such environments with ease, but today's deep learning systems struggle with them. Standard Artificial Neural Networks (ANNs) often fail dramatically when learning multiple tasks, a phenomenon known as {\it catastrophic forgetting} \citep{mccloskey1989, french1999} where the network forgets previously learned information. ANNs are inherently designed for static environments with batch training, and learning multiple sequential tasks can lead to significant interference between tasks. Embodied systems, where an agent actively behaves in a changing environment, pose additional challenges. In dynamic scenarios, the training dataset itself is not fixed. Sensory inputs are dependent on an agent's actions and as an embodied agent learns, the actions taken for a given context change as well. Thus, a network learning in these situations needs to avoid forgetting relevant information, update only the information that requires fine tuning, and forget the information that is no longer relevant. The network must distinguish between these types of information categories instead of treating all information as equivalent. The optimal algorithms and architectures for learning in dynamic environments are unknown and remain a fundamental research challenge for AI.

\begin{figure}[t]
\begin{center}
\begin{minipage}{.45\textwidth}
  \centering
  \includegraphics[width=2in]{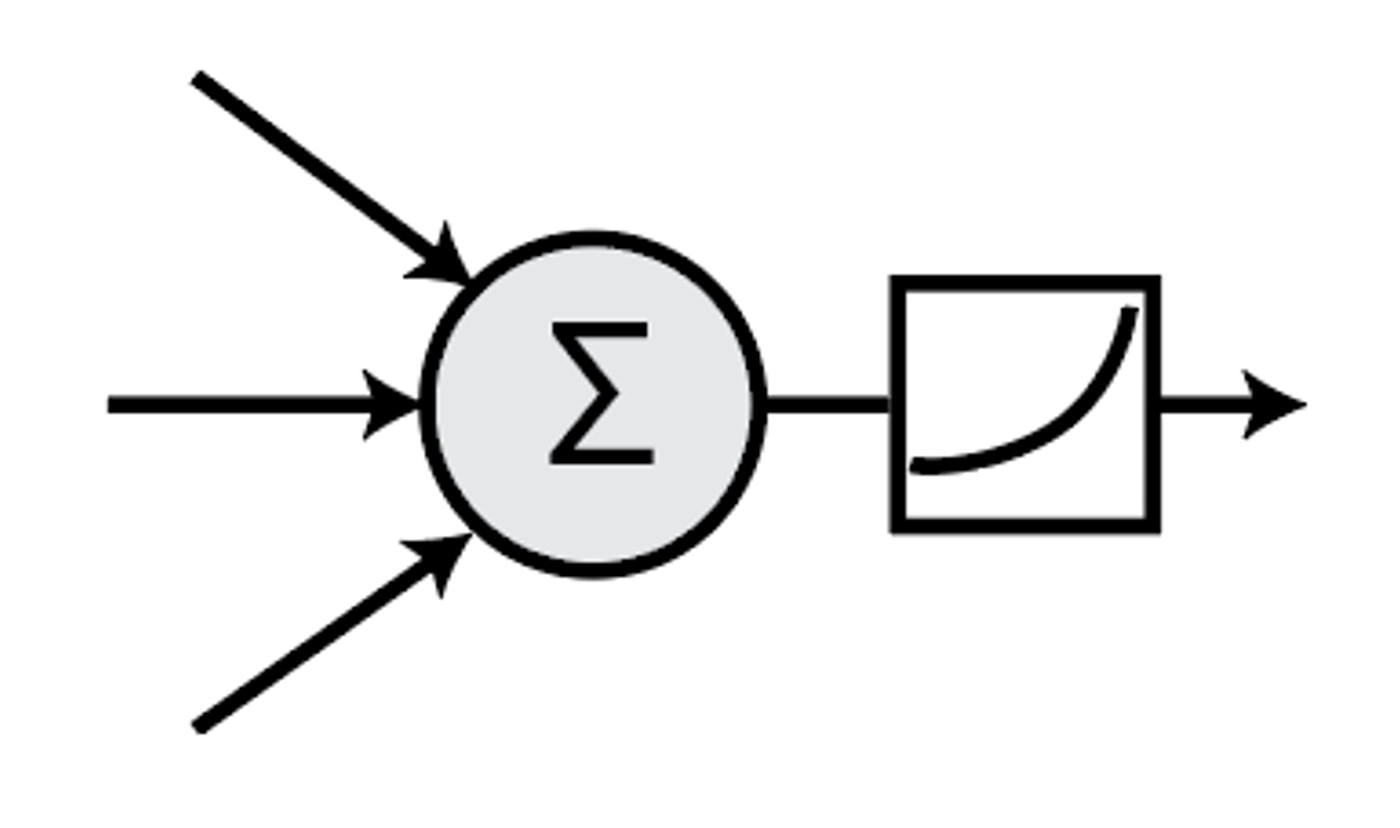}
\end{minipage}
\begin{minipage}{.45\textwidth}
  \centering
  \includegraphics[width=1.3in]{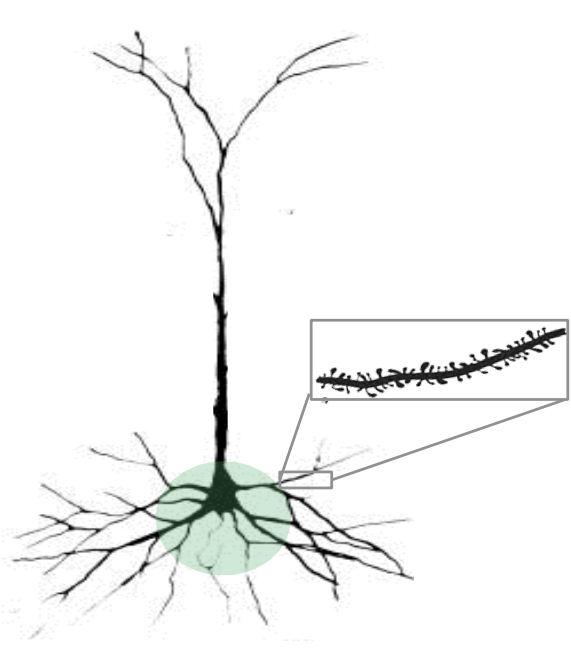}
\end{minipage}
\end{center}
\caption{
{\bf Left:} the point neuron prevalent in most ANNs today simply computes a linear weighted sum of its inputs followed by a non-linearity.
{\bf Right:} Morphology of a representative pyramidal neuron. Pyramidal cells in the brain exhibit a vastly more complex structure and functionality. Inset shows a prototypical basal dendritic segment that acts as an independent computational unit.
}
\label{fig:pyramidal}
\end{figure}

We investigate these questions by looking to neuroscience and biological systems for clues to inform ANNs. In particular we hypothesize that biological properties of {\it pyramidal neurons} in the neocortex can enable targeted context-specific representations that avoid interference. Most ANNs today rely on an idealized (but inaccurate) model of neurons known as the {\it point neuron model}, consisting of a linear weighted sum of inputs followed by a non-linearity (Figure~\ref{fig:pyramidal} left). Proposed over a hundred years ago \citep{lapicque1907}, the point neuron model continues to form the basis for current deep learning systems \citep{mcclelland1986parallel, lecun_deep_2015}. In contrast, pyramidal neurons, which comprise most cells in the neocortex, are significantly more sophisticated and demonstrate a wide range of complex non-linear dendrite-specific integrative properties \citep{spruston2008} (Figure~\ref{fig:pyramidal}, right). Experimental evidence suggests that dendrites are important for learning task-specific patterns \citep{yang2014}. In this article we incorporate into an ANN two properties of biological neural networks: active dendrites, and sparsity via local inhibition.

We explore the impact of these properties in two non-traditional machine learning scenarios: {\it multi-task reinforcement learning} (multi-task RL) and {\it continual learning}. In multi-task RL, a robotic agent learns to perform a diverse set of independent tasks \citep{yu2019}. Even though tasks are interleaved through training, standard ANNs suffer from significant task interference. In continual learning, a network is trained sequentially on a set of tasks and evaluated on all tasks after training \citep{mccloskey1989, van_de_ven_three_2019}. Here, standard ANNs do not perform well due to catastrophic forgetting. Specifically, because ANNs with point neurons overwrite most of their connections during each iteration of learning, tasks learned at the beginning of training are forgotten and receive low accuracy scores during the evaluation phase \citep{french1999, parisi_continual_2019}.

The rest of the article is arranged as follows. After discussing background material, we propose a new architecture that incorporates dendrites and sparse representations into deep learning. We then test our architecture on one representative benchmark from each of the two scenarios, multi-task RL and continual learning. We show experimental results on a standard multi-task RL benchmark, Meta-World. We also show results on a standard continual learning benchmark, permutedMNIST. The results in both cases show that an identical architecture with active dendrites performs well in both benchmarks. Finally, we analyze the results and show that active dendrites and sparse representations help with catastrophic forgetting and gradient interference by learning to create task-specific subnetworks where representations are sparse and mostly orthogonal. Overall, our results suggest that detailed biological properties of neurons can be used to address dynamic scenarios that are difficult for traditional ANNs to solve.

\section{Background}

\subsection{Multi-Task Learning}

The goal of multi-task learning \citep{caruana1997} is to learn a single function that can solve a variety of different learning tasks. The literature in multi-task learning spans many subfields of machine learning, including computer vision \citep{misra2016crossstitch, liu2019endtoend, purushwalkam2019task, kendall2019multi}, and natural language processing \citep{dong2015multi, mccann2018natural}.
The fields of multi-task RL and continual learning can be seen as subsets of multi-task learning. In the former, tasks are learned in parallel.
Conversely, in continual learning, tasks are learned in an ordered sequence.

Compared to single-task machine learning, learning multiple distinct tasks introduces new challenges. When using gradient-based learning algorithms such as backpropagation\footnote{In this paper the term \emph{backpropagation} refers to the learning method used in deep learning \citep{rumelhart1986} and not the phenomenon of back propagating action potentials in dendrites.}, one challenge is that error gradients and accumulated knowledge from different tasks can interfere with one another. The weight changes necessary to reduce the error for one task may be very different from the changes required for another task. This is a common problem sometimes defined as catastrophic forgetting \citep{french1999} or catastrophic interference \citep{mccloskey1989} in continual learning.

\citet{yu2020gradient} propose a method to modify conflicting gradients through gradient projection. Several other works demonstrate that using or changing the gradients via various normalization, gradient-similarity, and regularization techniques can improve learning in multi-task settings \citep{zhang2014regularization, chen2018gradnorm, sener2019multitask, du2020adapting}. Novel network architectures are an alternate strategy for avoiding interference in multi-task computer vision settings. \citet{rosenbaum2018routing} implement routing networks, learned functions that use task information to determine how to compose a set of function blocks. \citet{liu2019endtoend, maninis2019attentive} demonstrate that attention-based architectures could also prevent task interference in multi-task learning scenarios.

\subsubsection{Multi-Task Reinforcement Learning}
\emph{Reinforcement learning} (RL) is a branch of machine learning in which an agent acts in an environment and receives rewards for each action taken \citep{sutton2018reinforcement}. The goal is to train an agent, whose actions are determined by a \emph{policy function}, to maximize the total reward received. One fundamental challenge of RL is that the training set itself is highly dynamic. As the agent learns and updates its policy function, it chooses different actions, which in turn changes the sequence of inputs that are received.

\emph{Deep RL} uses deep learning networks to represent the policy function (see \citet{arullkumaran2017} for a review). Recent years have witnessed the promise of deep RL in a variety of different settings. \citet{mnih2013playing} demonstrate that an agent trained with their Deep Q-Network can surpass the performance of expert humans in Atari video games. A few years later, \citet{silver2018general} achieve superhuman performance in more challenging games such as Chess and Go. Other algorithms achieve strong performance in continuous environments with continuous action inputs \citep{lillicrap2016continuous}. Other methods attempt to induce beneficial learning behaviors such as more stable training \citep{schulman2017proximal} and improved exploration \citep{haarnoja2018soft}.

\emph{Multi-task reinforcement learning} combines Deep RL with multi-task learning \citep{wilson2007multi, yang2020multi, yu2020gradient}. Multi-task RL leads to particularly challenging and interesting scenarios where the system must address both dynamic training regimes and interference from multiple tasks. The idea of separating a neural network into different modules which are composed in a task-dependent manner is proposed in multi-task RL to prevent gradient interference \citep{devin2017learning, andreas2017modular, sahni2017learning, haarnoja2018composable, goyal2020reinforcement, yang2020multi}. Policy distillation, in which information from a ``teacher'' network is condensed to a smaller ``student network'', is another popular approach to combine information from different tasks in an effective manner \citep{rusu2016policy}.

\subsubsection{Continual Learning}
\label{section:continual_learning}

While multi-task RL requires the simultaneous acquisition of multiple skills, continual learning requires the sequential acquisition of multiple skills. More generally, continual learning is the ability to acquire new knowledge over time while retaining relevant information from the past.
A typical scenario involves training a network on a set of distinct tasks presented in a strict sequence of training phases. Testing the network involves measuring accuracy on all past tasks.
\citet{van_de_ven_three_2019} and \citet{parisi_continual_2019} extensively review the field.
Two common approaches to catastrophic forgetting in continual learning involve regularization and subnetworks methods.

Regularization-based methods in continual learning regulate plasticity levels throughout the network during the course of training.
In recent years, two of the most prominent examples of regularization are Elastic Weight Consolidation (EWC) \citep{kirkpatrick2017} and Synaptic Intelligence (SI) \citep{zenke2017}. Both methods (EWC and SI) estimate the relevance of each weight of the network in solving each task. Inspired by the complex synapse structures seen in biology, SI uses an additional parameter per weight with internal dynamics that depend on the relevance of each weight to each task.

Subnetwork-based methods reduce task interference by identifying subpopulations of neurons that each learn one of the many tasks in the sequence. Gated Linear Networks (GLNs) \citep{veness2021} and Dendritic Gated Networks (DGNs) \citep{sezener2021} are examples of this type of approach and work by applying a gating mechanism that selects subnetworks based on the input.
Context-dependent Gating (XdG) \citep{masse2018} selects predetermined subnetworks of neurons, but exact task information must be provided both at training and test times.  Similarly, in \citet{wortsman2020} each task is designated a sparse subset of neurons in the network.

\subsection{Properties of Biological Neurons}

Biological neural networks have evolved in ways that make them much more resilient to catastrophic forgetting and are able to perform significantly better in dynamical scenarios than any ANN to date.
ANNs and their component point neurons emerged as simplified abstractions of the complex processes occurring in biological networks and neurons respectively.
In this section, we explore the complexities of biological neural networks and review a few properties that are relevant to our work.

\subsubsection{Neurons and Active Dendrites}
\label{section:active_dendrites}

The \emph{pyramidal neuron} is the most prevalent neuron type found in the neocortex and hippocampal areas \citep{spruston2008,Ramaswamy2015}. In particular they represent the most common excitatory neuron type found in areas associated with advanced cognitive functions \citep{spruston2008}. A typical pyramidal neuron has an extensive dendritic tree containing thousands of synapses, each receiving input from another neuron \citep{bentivoglio_fine_2001, cajal1894, kandel_principles_2012}. The \emph{point neuron model} \citep{lapicque1907} postulates that all of these synapses have a linear impact on the cell. This simple assumption formed the basis for Rosenblatt's original Perceptron \citep{rosenblatt1958} and continues to form the basis for current deep learning systems and ANNs \citep{mcclelland1986parallel, lecun_deep_2015}.

Today it is well known that the point neuron assumption is an oversimplified model of biological computations.
Proximal synapses (close to the cell body) have a linear impact on the neuron, but the vast majority of synapses are located on distal dendritic segments (far from the cell body) and individually have minimal impact on the cell. These distal segments process groups of synapses locally in a non-linear fashion, and are known as \emph{active dendrites} \citep{Magee2000, antic2010, major2013, Stuart2015, Stuart2016}. Empirical evidence \citep{london_dendritic_2005, branco_single_2010} suggests that each distal dendritic segment acts as a separate active subunit performing its own local computation. Modeling studies show that neurons with active dendrites are more powerful and complex than the point neuron model can accommodate \citep{poirazi2003, poirazi2020, jadi2014, beniaguev2021}.

When input to an active dendritic segment reaches a threshold, the segment initiates a \emph{dendritic spike} \citep{antic2010}. In basal dendritic segments, dendritic spikes travel to the cell body and can depolarize the neuron for an extended period of time, sometimes as long as half a second \citep{antic2010, major2013, gao2021}. During this time, the cell is significantly closer to its firing threshold and any new input is more likely to make the cell fire. This suggests that basal active dendrites have a {\it modulatory}, long-lasting impact on the cell's response, with a very different role than proximal, or feedforward, inputs \citep{hawkins2016, antic2018}. Active dendritic segments typically receive contextual input that is a different input than received in proximal segments. These context signals can arrive from other neurons in the same layer, neurons in other layers, or in the form of top-down feedback. Recent experimental evidence has shown that the input on active segments can drive context-dependent activity \citep{takahashi2020}. In our model, we incorporate these ideas and explore the possibility of using context to create task-specific subnetworks.

\subsubsection{Sparse Representations}

Neural circuits in the neocortex are highly sparse.
Studies show that relatively few neurons spike in response to a sensory stimulus across multiple sensory modalities \citep{attwell2001, barth2012, liang_sparse_2019}.
Sparsity is also present in neural connectivity; cortical pyramidal neurons show sparse connectivity to each other and receive relatively few excitatory inputs from most surrounding neurons \citep{holmgren2003}.
These two phenomena are significantly different from standard ANNs, where both activations and connectivity are dense.

When modeling sparsity in ANNs, sparse neural representations are translated into vectors where most of the entries are off (i.e., equal to zero) \citep{majani1989}.
Just like in dense representations, individual entries in a sparse representation can correspond to the presence of certain features (e.g., the unique position of an edge in an input image).  One advantage of sparsity in representations is that vectors for two separate entities have low {\it overlap}, which means the set of features/entries that are non-zero in both vectors is small.
Previous studies show that sparse representations are more resistant to noise than dense representations, and slight perturbations in the input are less likely to hinder a trained pattern recognizer \citep{ahmad2016, ahmad2019, Paiton2020}.
The idea of low representation overlap among unrelated inputs may be particularly useful when an ANN is learning multiple, unrelated tasks.
If the representations of two different tasks have near-zero overlap, it is possible to significantly reduce task interference. We explore this question in our simulations below.

\section{Active Dendrites Network Model}
\label{section:model}

Our primary goal is to augment standard ANNs with the biological properties described above. The extensions should be general and applicable to a range of complex scenarios such as multi-task RL and continual learning. The key aspects of our model are summarized as follows, with details noted in the rest of this section:

\begin{enumerate}
\item Pyramidal neurons integrate a range of diverse inputs on multiple independent dendritic segments. To model this, we implement neurons that separate out contextual inputs from  feedforward inputs. Each neuron processes the feedforward input using a linear weighted sum. A set of independent dendritic segments process the contextual input using a separate set of weights.
\item Contextual inputs on active dendrites can modulate a neuron's response, making it more likely to fire. To model this, we implement a function that can up-modulate or down-modulate the feedforward activation based on dendritic activation.
\item Neural activity and connectivity are highly sparse. To model this, we incorporate a $k$-Winner-Take-All function ($k$WTA) that mimics biological inhibitory networks \citep{Cui2017} and guarantees sparse activations.
\end{enumerate}

\noindent The above properties are implemented such that the entire network is differentiable and trainable end-to-end using backpropagation. This makes the architecture suitable for testing on any standard deep learning scenario.

\subsection{Active Dendrites Neuron}
\label{section:neuron-model}

\begin{figure}[t]
\centering
\includegraphics[width=6.5in]{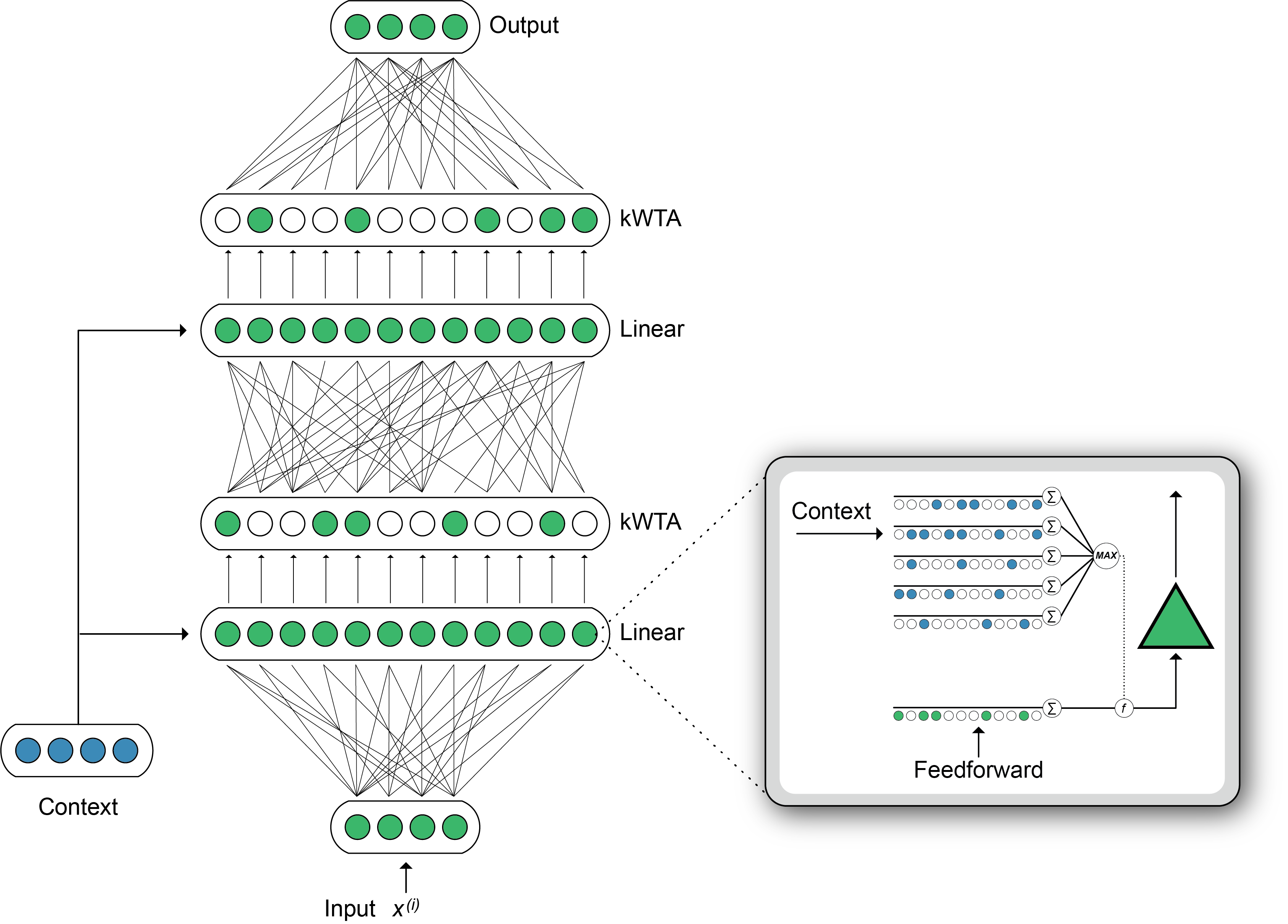}
\caption{
{\bf Right (inset):} Illustration of a single Active Dendrites Neuron.
Feedforward weights (green) receive regular feedforward input while dendritic segments (blue) receive a context vector.
After all dendritic segments compute an activation value, the highest value modifies the linear weighted sum computed by feedforward weights.  {\bf Left: }An overview of the base network structure used in our experiments.
There are 2 hidden layers. Each layer outputs sparse activations, as determined by a $k$WTA activation function. In addition, the weights between layers can be sparse. A context vector is computed for each input. The dendritic segments in each layer receive this context vector as input.
}
\label{fig:base_network_with_neuron}
\end{figure}

Building on the original HTM neuron model \citep{hawkins2016}, our \emph{Active Dendrites Neuron} (Figure~\ref{fig:base_network_with_neuron} (right inset)) receives two sources of input, analogous to the proximal and distal inputs in pyramidal neurons. Feedforward activation is computed by a linear weighted sum of the feedforward input vector, identical to the mechanism in a point neuron. Meanwhile, multiple dendritic segments process a context vector, and the subsequent dendritic output modulates the feedforward activation. This computation produces a neuron where the magnitude of the response to a given stimulus is highly context-dependent. 

Given input vector $\x$, weights $\w$, and bias $b$, our neuron computes the following feedforward activation:
\begin{align} \label{eq:ff output}
    \hat{t} &= \w^\top \x + b
\end{align}

\noindent Similarly, each dendritic segment $j$ computes $\u_j^\top \context$, given weight $\u_j$ and context vector $\context$. (The method we use to compute the context vector, $\context$, is described in later sections.) We select the segment with the strongest response to the context when computing dendritic activation $d$, which is used to modulate the neuron:
\begin{align} \label{eq:dendrite activation}
    d = \max_j \u_j^\top \context
\end{align}

In order to modulate feedforward activation $\hat{t}$ by the dendritic activation $d$, we use modulation function $f(\hat{t}, d)$ where $f(m, n) = m \times \sigma(n)$. Here, $\sigma(\cdot)$ is the sigmoid function which takes a real number and maps it into the range $[0, 1]$. Therefore, by combining~\eqref{eq:ff output} and~\eqref{eq:dendrite activation} with $f$, we can write the output of a single Active Dendrites Neuron as:
\begin{align}
    \hat{y} &= f\left(\hat{t}, d \right) \\
    &= f\left(\w^\top \x + b, \max_j \u_j^\top \context \right) \\
    &= \left( \w^\top \x + b \right) \times \sigma \left( \max_j \u_j^\top \context \right)
\end{align}

\noindent Here, a strong positive dendrite response to the context vector will retain the feedforward activation. Conversely, weak or negative responses to the context vector will significantly reduce the activation. We note that there are many variations of~\eqref{eq:dendrite activation} that are possible. We found that the network works best when we select the dendrite activation with the largest absolute value and retain the sign in $d$ (Section~\ref{section:absmax}).

\newpage

\subsection{Sparse Representations}

We apply a $k$WTA activation function \citep{ahmad2019} as our choice of non-linear activation in each hidden layer of the network:

\begin{equation}
    k(\hat{y}_i) = 
    \begin{cases}
        \hat{y}_i & \text{if } \hat{y}_i \text{ is one of the top } k \text{ activations over all } i \\
        0 & \text{otherwise}
    \end{cases}
\end{equation}

\noindent where $i$ indexes neurons in the same layer. The effect of $k$WTA is to ensure sparsity by selecting the top $k$ activations and setting all others to zero. Feedforward layers that are modulated by dendritic segments and apply $k$WTA thus produce sparse activity patterns that are highly context-dependent. Additionally, our feedforward layers also use sparse weights as proposed in \citet{ahmad2019}.

\subsection{Active Dendrites Network Architecture}

\vspace{-1mm}

Figure~\ref{fig:base_network_with_neuron} (left) shows our Active Dendrites Network, trained end-to-end with backpropagation, where all neurons in each hidden layer are Active Dendrites Neurons. We make two notes: first, only the neurons that were selected by the $k$WTA function will have non-zero activations (and thus non-zero gradients).
Therefore, during the backward pass, only the weights corresponding to those winning neurons will be updated.
Second, for each of those winner neurons, only the dendritic segment $j$ that was chosen by the $\max$ operator is updated; all other segments $\u_{j'}$ for $j' \neq j$ remain untouched.
Thus a very small sparse subset of the full network is actually updated for each input.

We hypothesize that a functional specialization will emerge where different dendritic segments will each learn to identify specific context vectors. Since most dendritic segments that don't respond to a specific context will not be updated, any context-dependent modulation of the neuron should be preserved from task to task. Ideally, the whole process will preserve any context-dependent modulation of a neuron between tasks, reduce gradient interference, and prevent catastrophic forgetting.

\section{Results}
\label{section:experiments}

\subsection{Results with Multi-Task Reinforcement Learning}

\begin{figure}[h]
\centering
\includegraphics[width=5in]{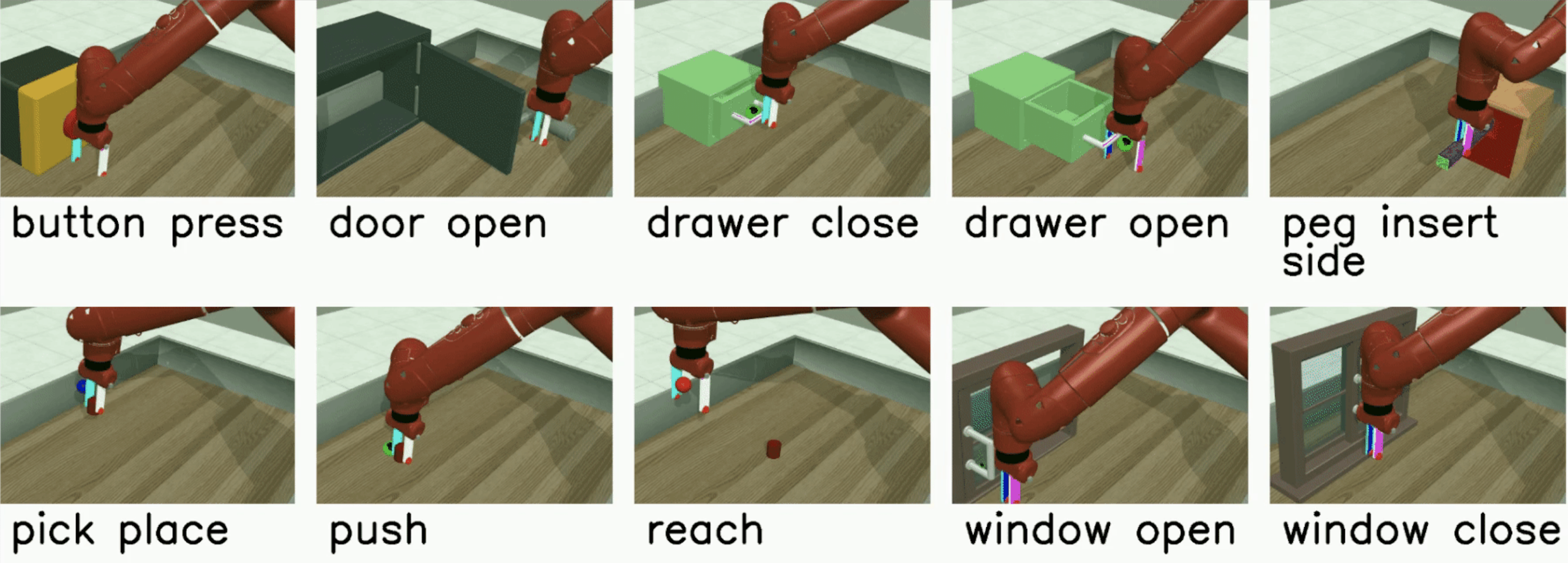}
\caption{
The Meta-World v2 Multi-Task 10 (MT10) environment, where a single robotic arm must learn to solve a variety of tasks ranging in difficulty.  
}
\label{fig:metaworld_env}
\end{figure}

The multi-task RL problem we investigate uses the Meta-World v2 environment and its associated v2 tasks \citep{yu2019}.
Meta-World contains multiple different object manipulation tasks that a single robotic arm must learn to solve simultaneously.
We use the MT10 environment, which contains 10 tasks ranging in complexity as depicted in Figure~\ref{fig:metaworld_env}.
Although the concrete outcome of each task is unique, all tasks share a common structure that enables the agent to leverage some shared information during training.
For instance, learning how to grasp an object is a shared concept among many of the tasks.

The algorithm we use to train our robotic agent is multi-task Soft Actor-Critic (MTSAC) as introduced by \citet{yu2019}, an adaptation of the popular Soft Actor-Critic (SAC) framework proposed earlier by \citet{haarnoja2018soft}.
MTSAC is an actor-critic deep RL algorithm that maximizes an agent's cumulative reward to solve a task while also maximizing entropy to encourage environment exploration. To maintain consistency with the codebase of \citet{yu2019} which fixes goal states (e.g., position of an object in the environment) through training, we also keep goal locations constant across all our experiments. 
A deeper explanation about our multi-task RL setup and the algorithm we use to train the agent can be found in Section~\ref{methods:RL}.
As in many RL problems, there is no static training and testing dataset.
Rather, past experiences from the agent are used to iteratively train the agent. We freeze the network at regular intervals to test accuracy on all tasks. 

\subsubsection{Network Structure for Multi-Task RL}

\begin{figure}[h]
\centering
\includegraphics[width=5in]{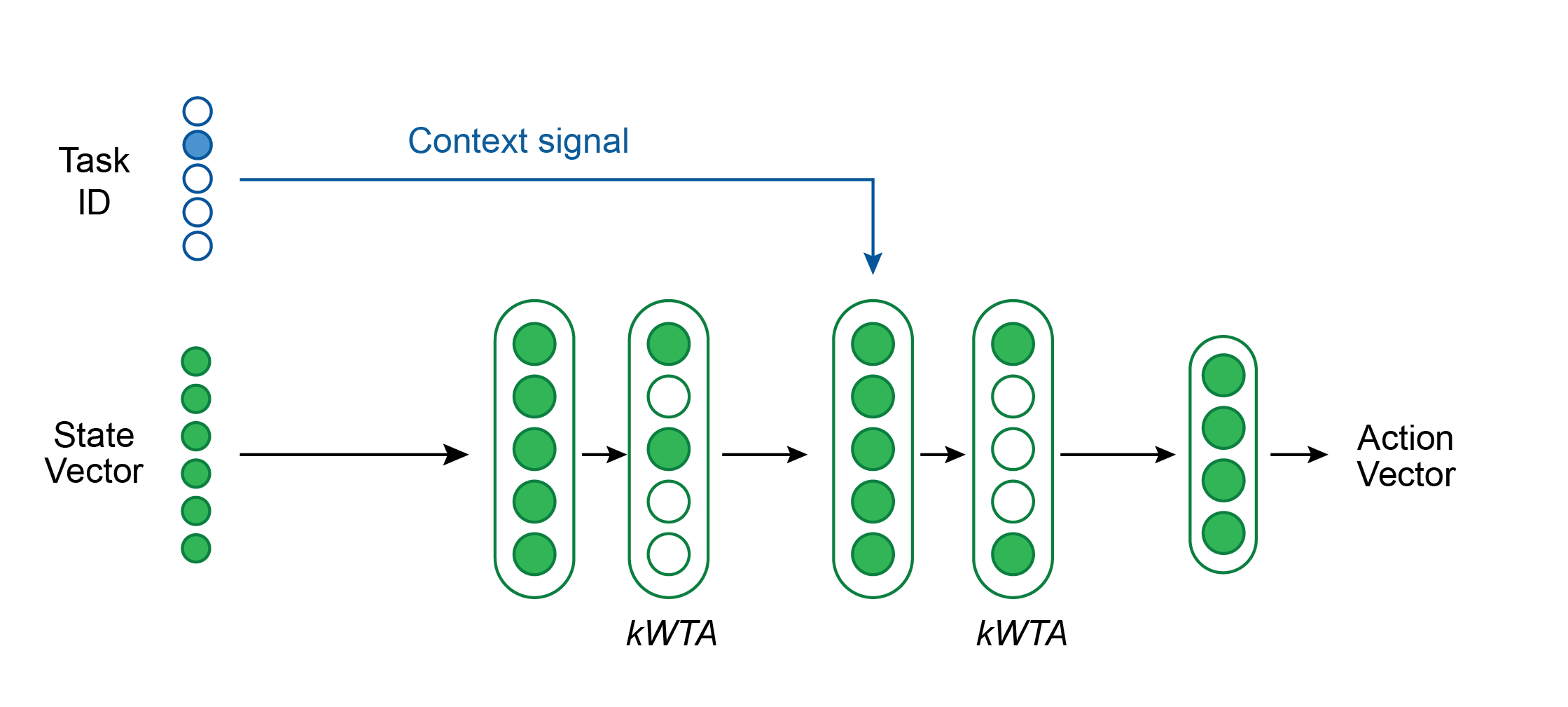}
\caption{
An overview of the network structure used in our multi-task RL experiments.
A $k$WTA activation is applied to both hidden layers.
The context vector is the task ID.
The dendritic segments in the second hidden layer receive this context vector as input.
}
\label{fig:mtrl_network}
\end{figure}

Figure~\ref{fig:mtrl_network} shows our network architecture for multi-task RL.
We use a network with 2 hidden layers---each with 2,800 neurons and followed by a $k$WTA activation function---and a final output layer.
The first hidden layer has standard neurons whereas the second hidden layer contains Active Dendrites Neurons which are modulated by the context vector.
The primary feedforward input to the network is a state vector consisting of the agent's position in the world as well as the position and orientation of the target object.
The output of the network is an action vector that describes the joint torques and gripper forces of the robotic arm.
The structure of the state and output vectors is identical across all tasks.
All feedforward weights are sparse.

For our multi-task RL experiments, the context vector $\context$ encodes the task ID as a one-hot encoded vector.
We considered other options to generate $\context$, such as first pre-processing the one-hot encoding by a linear layer, but found that a one-hot encoding was adequate.
Each Active Dendrites Neuron in our network has exactly 10 dendritic segments (same as the number of tasks to learn) so that each segment can potentially learn to recognize a unique context vector.

We compare our Active Dendrites Network to baselines reconstructed\footnote{We are unable to directly present the published baseline results because their plots contain inconsistencies between success rates per-task and across all tasks. To present a fair comparison, we re-run the baseline networks using their codebase and hyperparameters.} from \citet{yu2019} which are standard multi-layer perceptrons (MLPs) with dense weights and ReLU activations. These MLP baselines are used to model both the policy and the Q function. Additionally, these MLP baselines receive context information $\context$ in the form of feedforward input concatenated to the state vector.
Thus both the Active Dendrites Network and the baseline network receive \emph{identical} information at each time step; the primary difference between the two architectures is how the context vector is handled.

Table~\ref{table:mtrl_hyperparameters} in Section~\ref{methods:RL_experiment_settings} shows the networks we ran, the number of \emph{non-zero parameters} in each network, and the hyperparameters used to train each network.
Although we control the hidden sizes to yield approximately the same number of total non-zero parameters across our experiments, we note that the MLP baseline network contains nearly 500,000 more non-zero parameters than our Active Dendrites Network.
We chose a network with 2 hidden layers to draw fair comparisons with the MLP baselines presented in \citet{yu2019}. Appendix~\ref{section:additional_mtrl_experiments} includes the results of additional experiments detailing the impact of some of our architectural choices. 

\subsubsection{Dendrites Improve Multi-Task RL Accuracy}

We show results from two different experiments that compare our Active Dendrites Network to the MLP baseline network.

\begin{figure}[h]
\begin{center}
\begin{minipage}{.49\textwidth}
  \centering
  \includegraphics[width=3.0in]{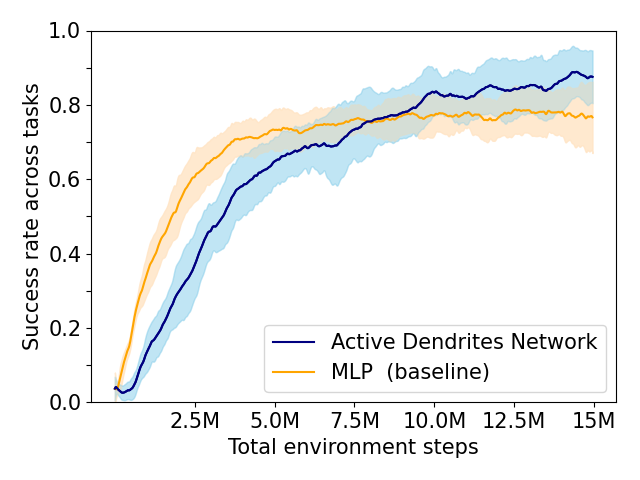}
\end{minipage}
\begin{minipage}{.49\textwidth}
  \centering
  \includegraphics[width=3.0in]{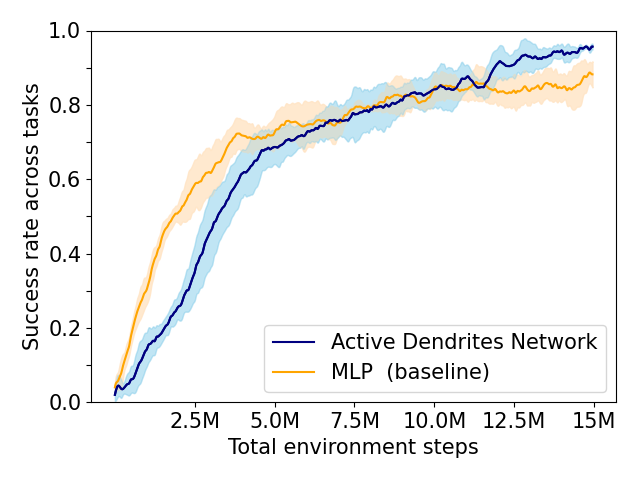}
\end{minipage}
\end{center}
\caption{
The success rate of our network when learning 10 tasks compared to the MLP baseline with context.
{\bf Left:} \textit{Experiment 1} - the average of 10 Active Dendrites Network runs and 10 MLP baseline network runs that all share the same training hyperparameters.
{\bf Right:} \textit{Experiment 2} - the average of the 5 best Active Dendrites Network experiments and the 5 best MLP baseline experiments.
The shaded region in each plot represents the standard deviation of the success rate from the average.
}
\label{fig:mtrl_accuracy}
\end{figure}

\begin{figure}[h]
\centering
\includegraphics[width=4.2in]{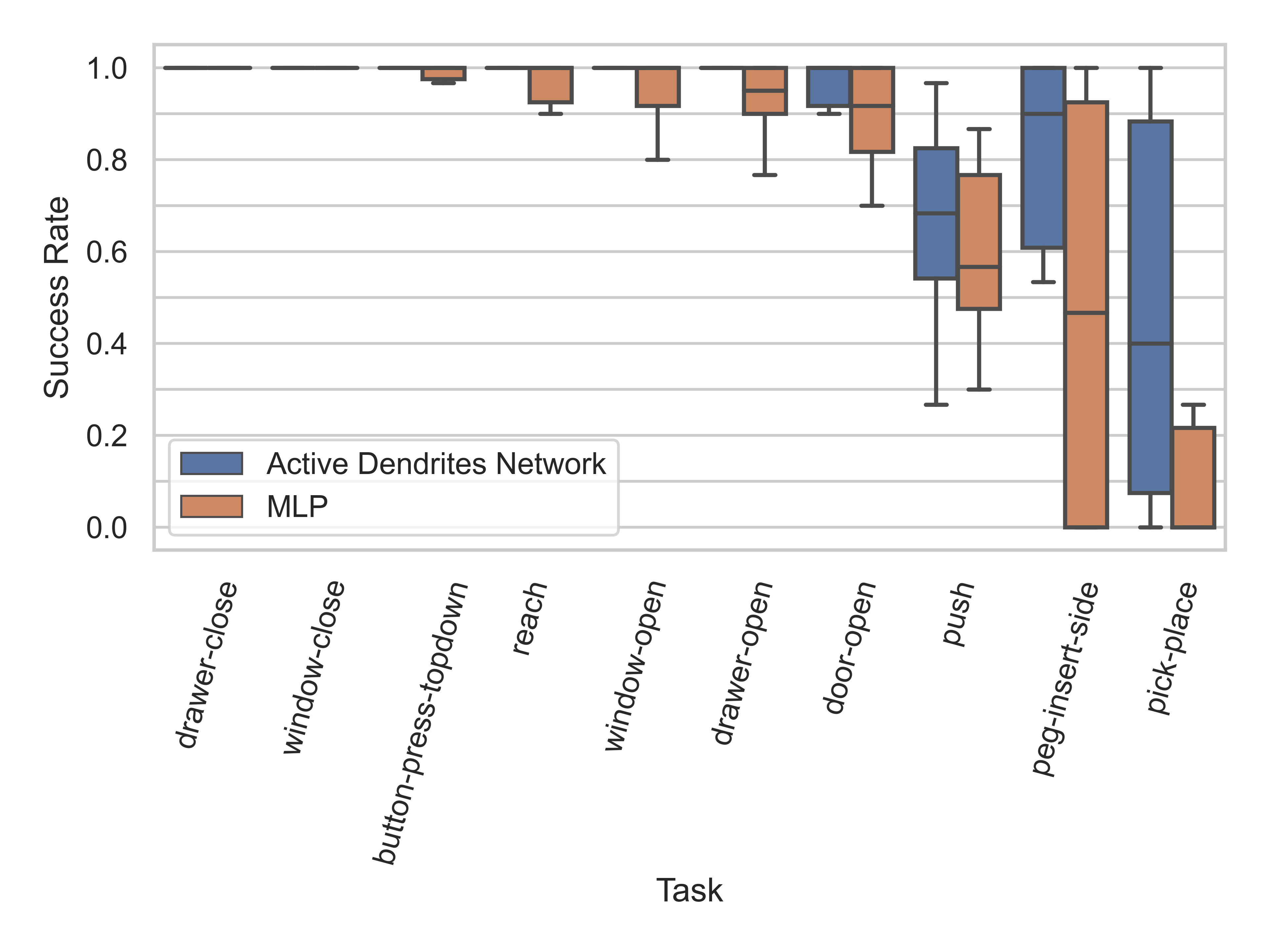}
\caption{
Box plots of the accuracies for each MT10 task for our Active Dendrites Networks and MLP baseline networks in Experiment 1. We discard outliers for all runs for clarity.}
\label{fig:mtrl_accuracy_per_task}
\end{figure}

\noindent \underline{\textit{Experiment 1}}: In this experiment, we assess the overall performance of each architecture.
We ran both an Active Dendrites Network and a MLP baseline network with identical training hyperparameters.
Figure~\ref{fig:mtrl_accuracy} (left) shows the mean overall success rate for each architecture during the course of training over 10 independent trials.
To identify which architecture performs the best for each task, we compute the mean success rate per task for the last 500,000 environment steps of training and list these values in Table~\ref{table:mtrl_task_accuracy}.
Additionally, we show the per-task training statistics during this same segment of training, as seen in Figure~\ref{fig:mtrl_accuracy_per_task}.

We see in Figure~\ref{fig:mtrl_accuracy} (left) that although the Active Dendrites Network has lower success rates early in training, it overtakes the baseline architecture and is about \acc{10} better by the end.
Table~\ref{table:mtrl_task_accuracy} shows that the average end success rate for the Active Dendrites Network (across the last 500,000 steps of training) is \bacc{87.5}.
In comparison, the average success rate for the MLP baseline is \bacc{76.6}.
We also note that the \textit{push}, \textit{peg-insert-side}, and \textit{pick-place} tasks were the hardest to solve because they are the most unlike the other tasks.
Specifically, these three tasks require that a robot grasp and move a small object to a specified location.
As evident in Figure~\ref{fig:mtrl_accuracy_per_task} for these three tasks, the median success rate of an Active Dendrites Network is far greater than that of the MLP baseline network.
We hypothesize that these tasks are hard to learn because of significant gradient interference with the other tasks, and that the context-specific sparsity imposed by the Active Dendrites Network helps remove this interference.

\begin{table*}[h]
\vskip 0.15in
\begin{center}
\begin{tabular}{lcc}
\toprule
\multicolumn{1}{l}{Tasks} & 
\multicolumn{2}{c}{Model} \\
\cmidrule(lr){2-3}

&
\multicolumn{1}{c}{Active Dendrites Network} &
\multicolumn{1}{c}{MLP Baseline} \\
\midrule
drawer-close & \bacc{100.0} & \bacc{100.0} \\
window-close & \bacc{99.7} & \acc{95.3} \\
button-press-topdown & \acc{95.7} & \bacc{97.3} \\
reach & \bacc{99.7} & \acc{86.3} \\
window-open & \bacc{99.3} & \acc{93.7} \\
drawer-open & \bacc{99.7} & \acc{86.0} \\
door-open & \bacc{94.7} & \acc{84.3} \\
push & \bacc{67.3} & \acc{59.0} \\
peg-insert-side & \bacc{71.7} & \acc{47.0} \\
pick-place & \bacc{47.7} & \acc{17.3} \\
\midrule 
Overall & \bacc{87.5} & \acc{76.6} \\
\bottomrule
\end{tabular}
\end{center}
\vskip -0.1in
\caption{The mean per-task success rate produced by each network in Experiment 1. The success rates are averaged over the last 500,000 steps of training.}
\label{table:mtrl_task_accuracy}
\end{table*}

\noindent \underline{\textit{Experiment 2}}: The high variance in Figure~\ref{fig:mtrl_accuracy} (left) is inherent in many RL scenarios \citep{Irpan2018, Ibarz2021}. This is in large part due to the highly stochastic and dynamic training process. For instance, small variations in the trained policy can result in large variations in the agent's behavior which significantly impacts the data collected during training. Additionally, a policy can generate different behaviors during training when sampling from its predicted action distribution.

To control for some of this variation in training, for each network initialization we select the best result across different training runs. For each of 5 different Active Dendrite Network initializations, we ran 5 training runs and picked the run with the highest end success rate across the last 500,000 environment steps of training. We then compute the mean overall success rate across these 5 best runs. We follow the same procedure for finding the 5 best MLP baseline networks and compare the results in Figure~\ref{fig:mtrl_accuracy} (right). 

We find that this process significantly reduces the variance and that the best Active Dendrites Networks still outperform the best MLP baseline networks. Across the 5 best Active Dendrites Network runs, the average overall end success rate is \bacc{95.6}. In comparison, across the 5 best MLP baseline runs, the average overall end success rate is \bacc{88.2}.

\subsection{Results With Continual Learning}
A typical continual learning problem consists of training a neural network on a discrete number of tasks in sequence. Once a network is trained on a particular task, it does not encounter that task during training again. The goal is to learn all the tasks in sequence without forgetting previously learned tasks. 

We use the permutedMNIST dataset \citep{goodfellow2014}, a common benchmark in continual learning where each task requires classifying images of handwritten digits (0--9) after a unique pixel-wise permutation has been applied. Since the data distribution of each task changes and because neural networks are generally not permutation-invariant, forgetting occurs. 

We use the original MNIST training dataset of $60,000$ images to construct the dataset for a single task. Since we train on $\mathcal{T}$ consecutive tasks, the network is trained on a total of $\mathcal{T} \times 60,000$ images. Once training is complete, the network accuracy is calculated using a test set consisting of all $\mathcal{T}$ permutations applied to the MNIST test dataset of $10,000$ images.

We train our model to learn up to 100 tasks in sequence. The network is tested at the end of training by computing accuracy on the test set for all tasks. When attempting to learn $\mathcal{T}$ consecutive tasks, the hidden neurons are equipped with $\mathcal{T}$ dendritic segments each to give it sufficient capacity to recognize a unique context vector for each task. We report accuracy numbers by averaging over 8 independent runs each with a randomly-picked seed. See Section~\ref{methods:continual learning} for additional details.

\subsubsection{Computing the Context Vector}
\label{section:context}

\vspace{-0.5mm}

\begin{figure}[t]
\centering
\includegraphics[width=6.7in]{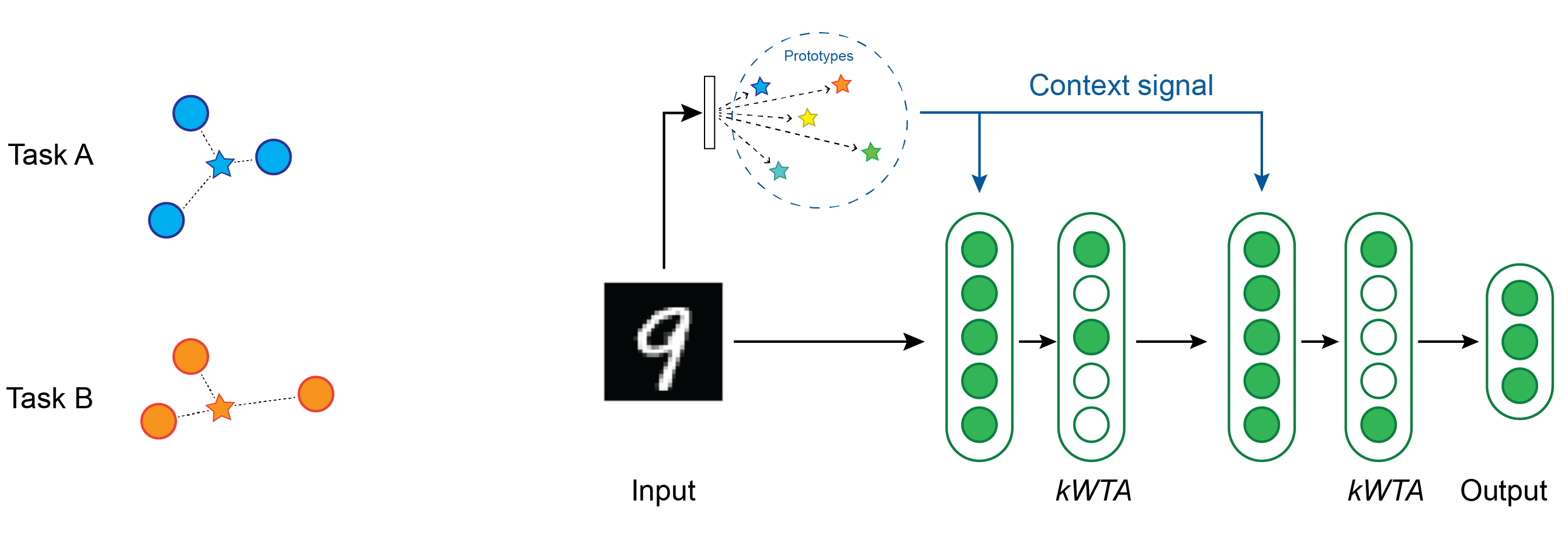}
\caption{
{\bf Left:} An illustration of the prototype method for computing the context vectors. The blue circles are training samples in input space for task A, while the orange circles are training samples for task B. The blue star is a vector that represents the prototype for task A, and the orange star represents the prototype for task B. {\bf Right:} An overview of the network structure used in our continual learning experiments. There are 2 layers of hidden units, each with a $k$WTA activation function. A context vector is computed from each image by locating the nearest prototype vector.
}
\label{fig:cl_network}
\end{figure}

As with multi-task RL, we need to compute an appropriate context vector.
For continual learning, we use a simple prototype method \citep{rosch1975, snell2017} to select the context vector where a single vector represents each task (Figure~\ref{fig:cl_network} (left)).
We implement two different variations of the prototype method depending on the knowledge available to the system during training.

\noindent {\bf Training method 1 (task information provided):} In the first method, we assume that the system receives task information during training, when all training samples for a particular task are assigned a single prototype context vector. We compute the prototype vector for task $\tau$ by taking the element-wise mean over all the training samples across all features: $$\p_\tau = \frac{1}{\left| V_\tau \right|} \sum_{\x \in V_\tau} \x$$ where $V_\tau$ denotes the set of all data samples $\x$ that the model observes to train on task $\tau$.
The dimensionality of the context vector is thus identical to the dimensionality of the input vectors. This context vector is specific to each task and agnostic to the target label.

\noindent {\bf Training method 2 (task information not provided):} In the second method, we relax the constraint that the identity of the task is given during training and instead implement prototypes that are automatically selected during training. To achieve this, we use a statistical clustering approach that builds context prototypes on the fly. When the system receives a new batch of training samples from a task, we use an unpaired multivariate $t$-test to compare the current samples to previously observed training samples.
If the new batch of samples is similar to earlier training samples, they are assigned to an existing prototype. If not, the new batch of samples is assumed to correspond to a new task, and a novel prototype is instantiated. In this case, there isn't necessarily a one-to-one mapping between tasks and prototype context vectors. More details on this method are described in Section~\ref{section:training_prototype}.

\noindent {\bf Selecting prototypes during inference:} For both methods above, we do not provide any task information to the system during evaluation. Instead it must dynamically select the correct context vector and provide that to the network. We enable this dynamic approach by selecting the closest prototype vector to each test example using Euclidean distance. That is, for a test example $\x'$, the chosen prototype is: $$\argmin_{\p_\tau} || \x' - \p_\tau ||_2$$ computed over all prototypes $\p_\tau$ stored in memory.

\subsubsection{Network Structure for Continual Learning}

Figure~\ref{fig:cl_network} (right) shows the network that we used for our continual learning experiments.
Each of the 2 hidden layers contain 2,048 Active Dendrites Neurons followed by a $k$WTA activation function.
The output of the network is a standard layer with 10 neurons.
We choose our network layer sizes to be similar to previous studies that report results on this dataset \citep{kirkpatrick2017, zenke2017, masse2018}.
(Section~\ref{methods:continual learning} details the hyperparameters used for each experiment.)

\subsubsection{Dendrites Mitigate Catastrophic Forgetting in Continual Learning}

\begin{figure}[h]
\begin{center}
\begin{minipage}{.49\textwidth}
  \centering
  \includegraphics[width=3.0in]{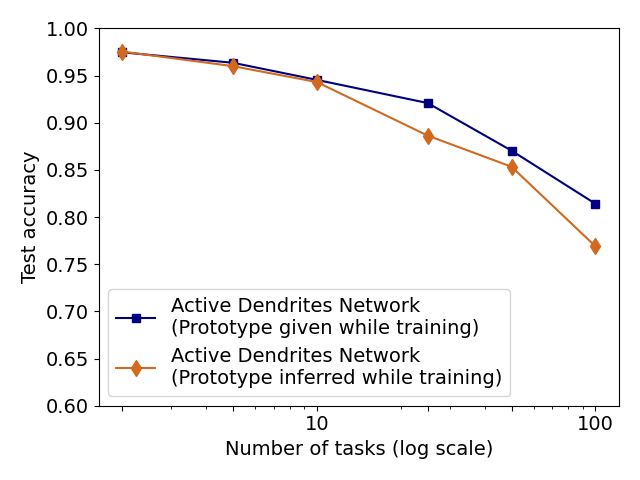}
\end{minipage}
\begin{minipage}{.49\textwidth}
  \centering
  \includegraphics[width=3.0in]{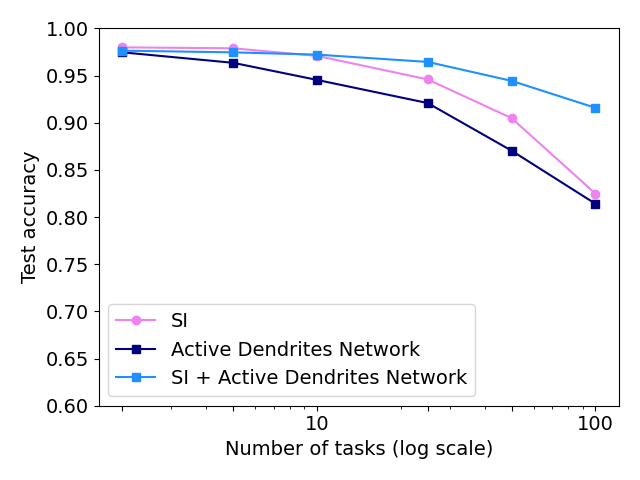}
\end{minipage}
\end{center}
\caption{
{\bf Left:} The accuracy of our Active Dendrites Networks when learning 2, 5, 10, 25, 50, and 100 permutedMNIST tasks in sequence. We show results using both prototype methods while training: when the the model is provided with a prototype, and when it must select the vector in an online manner. {\bf Right:} The accuracy of the Active Dendrites Network and SI. The accuracy when combining SI + active dendrites is greater than either one on its own.
}
\label{fig:si_accuracy}
\end{figure}

As shown in Figure~\ref{fig:si_accuracy} (left), we achieve accuracies of \bacc{94.6} and \bacc{81.4} on 10 and 100 consecutive permutedMNIST tasks, respectively, when context is provided during training, and accuracies of \bacc{94.3} and \bacc{76.9} when context is dynamically chosen during training.
Since there are always 10 categories, chance accuracy is \acc{10} independent of the number of tasks.
This demonstrates that the network successfully retains the majority of the knowledge from previous tasks.
Note that a standard feedforward network performs poorly on this benchmark \citep{kirkpatrick2017, zenke2017, van_de_ven_three_2019} (see also Section~\ref{section:dendrites_vs_more_layers} for more direct comparisons).

% Results with SI
We also compare the results with SI \citep{zenke2017} (see Section~\ref{section:continual_learning}).
SI is inspired by the complex structure of biological synapses and known to do well on this benchmark.
SI operates solely at the level of synapses: it maintains an  additional parameter per weight that controls the speed of weights adapting to specific tasks. In SI, the weight updates are sprinkled throughout the network and not grouped according to units or dendrites. On the other hand, the dendrites in our network impact a small subset of the neurons, and only the weights on these neurons and dendrites are modified. As such, our two approaches seem to be complementary.
Figure~\ref{fig:si_accuracy} (right) shows the benefits of combining these two techniques.
The accuracy of Active Dendrites Networks combined with SI improves to \bacc{97.2} and \bacc{91.6} accuracy on 10 and 100 consecutive tasks, respectively.
Combining the two leads to higher accuracy than either method on its own.
This suggests that biological mechanisms at the synapse, neuron, and network levels can operate together to handle continual learning.
Note that SI as described in \citet{zenke2017} requires knowledge of the task during training; therefore we only combine it with our first prototype method.
It may be possible to remove this restriction, which is a direction for future research.

\subsubsection{Comparison With Context Dependent Gating}
\label{section:xdg}

The idea of leveraging sparse representations and subnetworks within an ANN to combat catastrophic forgetting is not entirely novel.
The implementation closest to ours is XdG \citep{masse2018} that uses a hard-coded distinct subnetwork for each task. When training on a task, the implementation invokes the task-specific subset of the hidden layer of the ANN; other neurons are forced to have an activation value of zero. The XdG implementation requires a task ID that determines exactly which neurons to turn on or off. Training Active Dendrites Networks in a continual learning scenario also yields subnetworks and sparse representations. However, we emphasize two major distinctions between our model and XdG:
\begin{enumerate}
    \item Task information is inferred in our system (via prototyping) whereas XdG provides the system with a task ID during training and testing. As such, our system is solving a problem that is known to be significantly more challenging \citep{van_de_ven_three_2019}.
    \item Subnetworks automatically emerge via the use of dendritic segments for each new task whereas XdG pre-allocates a different subnetwork for each task, which also indicates our system is solving a more challenging problem.
\end{enumerate}

\begin{figure}[h]
\begin{center}
\begin{minipage}{.49\textwidth}
  \centering
  \includegraphics[width=3.0in]{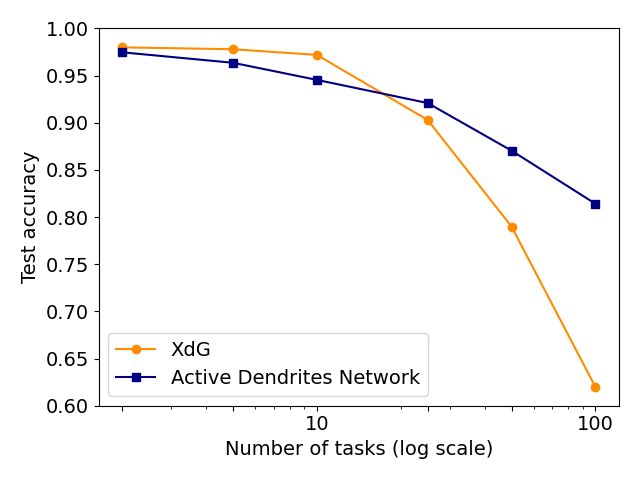}
\end{minipage}
\begin{minipage}{.49\textwidth}
  \centering
  \includegraphics[width=3.0in]{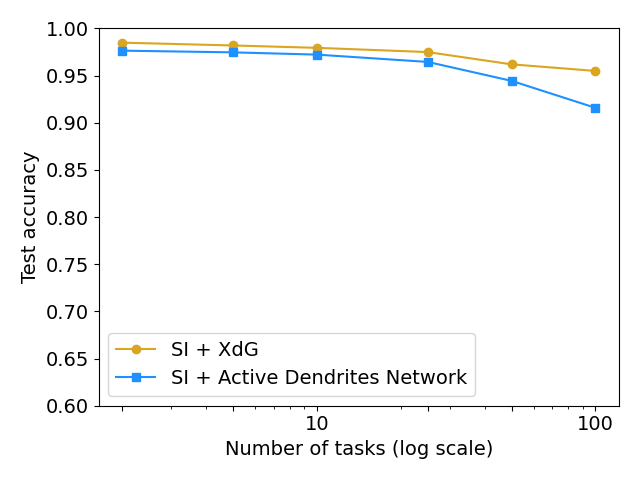}
\end{minipage}
\end{center}
\caption{
{\bf Left:} Final accuracy of the Active Dendrites Network in comparison to XdG when learning 2, 5, 10, 25, 50, and 100 permutedMNIST tasks. The more tasks learned by the system, the greater the accuracy of the Active Dendrites Network.
{\bf Right:} Final accuracy of each method when augmented with SI, and SI itself. XdG results are taken from \citet{masse2018}.
}
\label{fig:accuracy}
\end{figure}

We compare Active Dendrites Networks to XdG in Figure \ref{fig:accuracy}. Just as we augment Active Dendrites Networks with SI, so too does XdG. Our results with a large number of tasks are significantly better than XdG, and slightly worse than XdG combined with SI, but without their limitations.

Learning is more challenging in our system as dendritic segments must learn the mapping between context vectors and different subnetworks. In effect, sparse representations and minimally overlapping subnetworks emerge organically in our model. We note that perhaps this makes learning more effective as dendritic segments can choose subnetworks that overlap more for tasks that are more semantically related, thus requiring less network capacity.

\subsection{Analysis}

\subsubsection{Are Dendrites Invoking Subnetworks?}
\label{section:subnetwork-results}

The hypotheses of our work are two-fold. First, Active Dendrites Networks modulate an \textit{individual} neuron's activations for each task. Second, $k$WTA activations use this modulation to activate subnetworks that correspond to each task. To test these hypotheses, we train and analyze an Active Dendrites Network for $10$ tasks in multi-task RL and continual learning scenarios and investigate the representations of a layer of neurons modulated by dendritic segments. 

\begin{figure}[h]
\begin{center}

\begin{minipage}{.7\textwidth}
  \centering
  \includegraphics[width=1.8in]{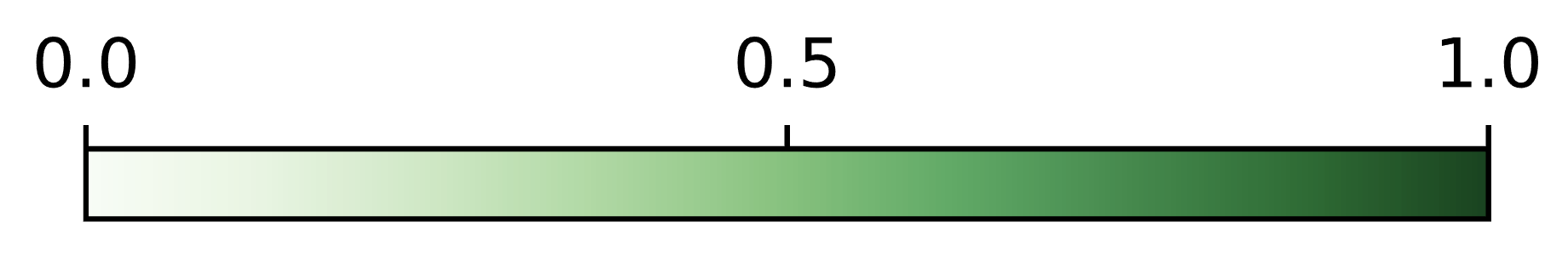}
\end{minipage}

\begin{minipage}{.49\textwidth}
  \centering
  \includegraphics[width=3.2in]{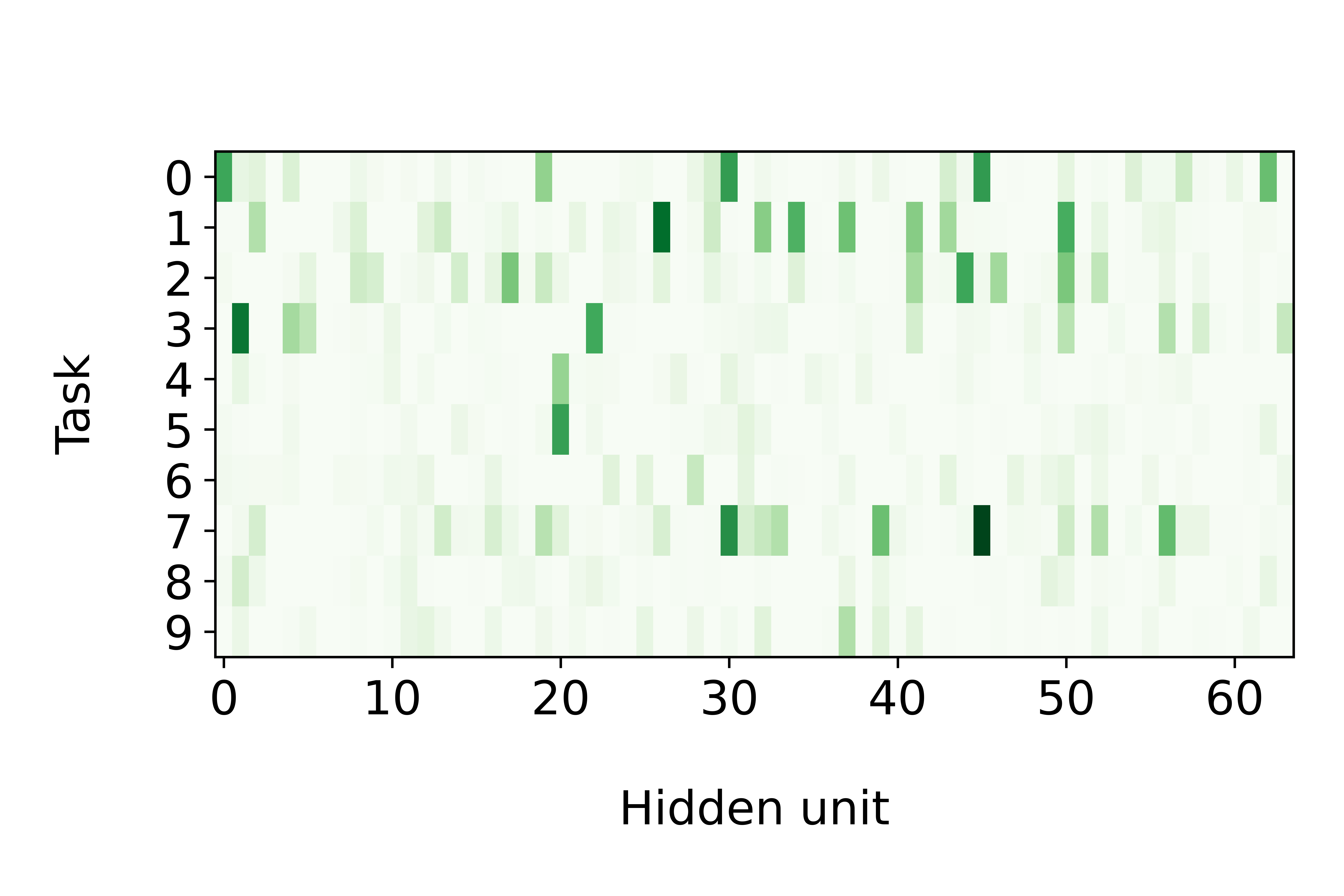}
\end{minipage}
\begin{minipage}{.49\textwidth}
  \centering
  \includegraphics[width=3.2in]{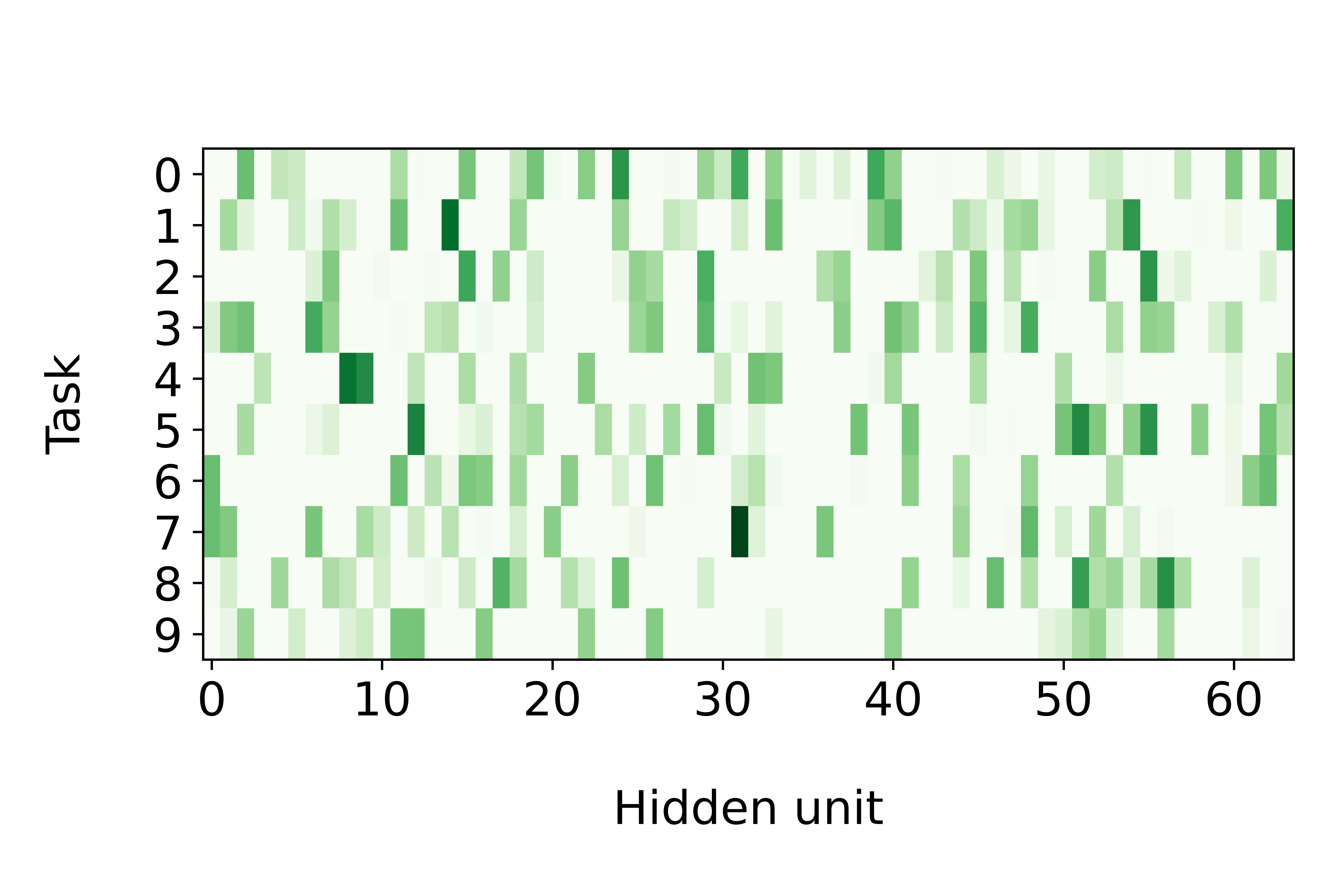}
\end{minipage}

\end{center}
\caption{
The fraction of instances for which each of the first 64 hidden units in the hidden layer became active (after applying $k$WTA), when training an Active Dendrites Network on MT10 tasks ({\bf left}) and 10 permutedMNIST continual learning tasks ({\bf right}).
Both figures separate instances by task.
For MT10, the figure tests the trained RL policy on each task three times during evaluation.
For permutedMNIST, the figure uses 5,000 randomly-chosen test examples across all tasks.
Note that each hidden layer contains more than 2,000 hidden units, but we show just 64 for ease of visualization.
}
\label{fig:activation_by_task}
\end{figure}

Figure~\ref{fig:activation_by_task} shows the average activation frequency per task (after applying $k$WTA) for the first $64$ neurons in the second hidden layer for both multi-task RL and continual learning. Looking horizontally across the rows, each task appears to select a different sparse subset of neurons. Looking vertically across the columns, each neuron appears to activate frequently only for a small fraction of tasks. According to this measure, it appears that the network has indeed learned to invoke minimally overlapping subnetworks for different tasks.

\begin{figure}[h]
\centering
\includegraphics[width=6.5in]{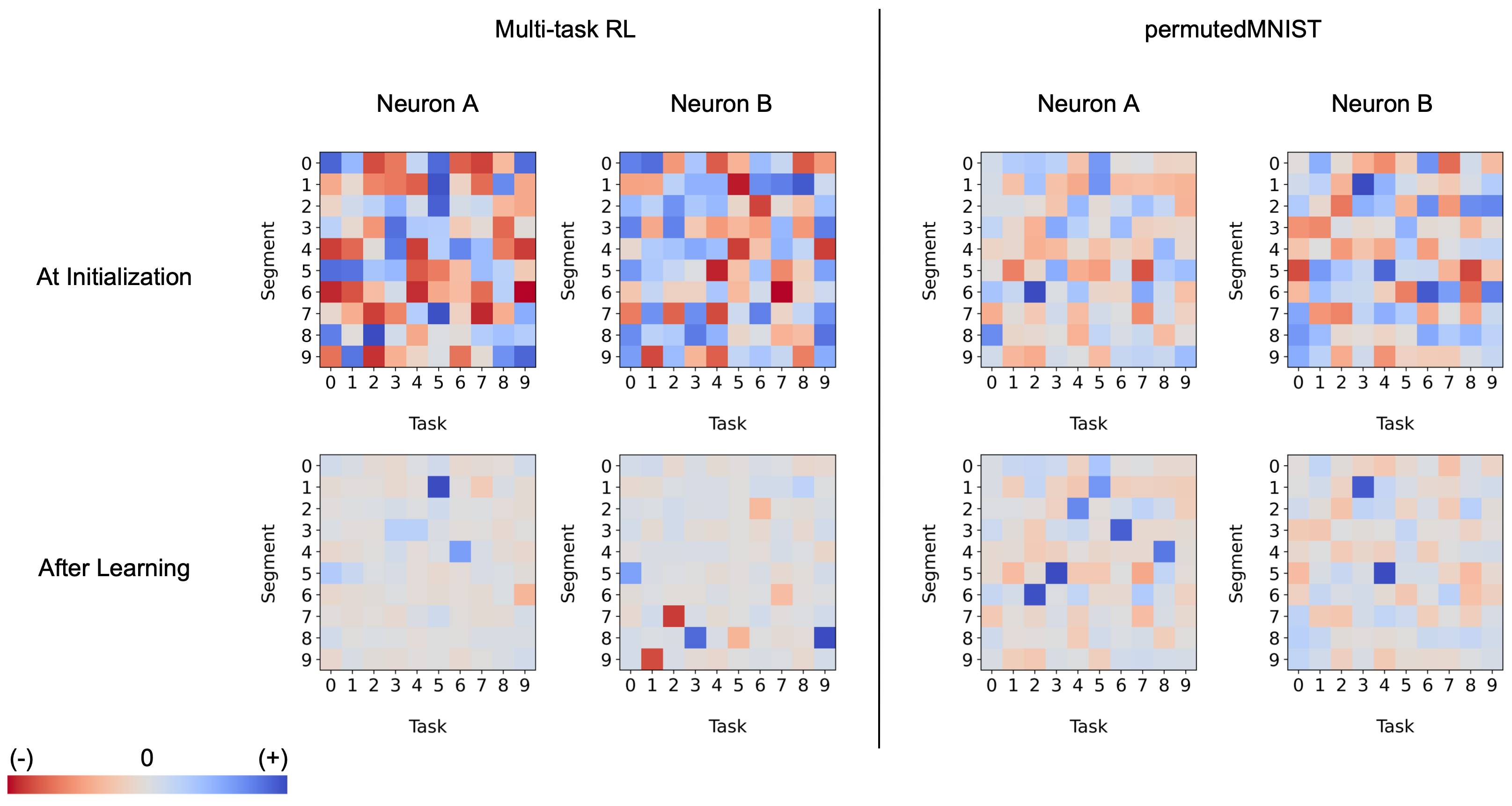}
\caption{
The behavior of the dendritic segments of two separate neurons in a hidden layer of an Active Dendrites Network during three random evaluations of each MT10 task and 5,000 random evaluations of each permutedMNIST task. These charts show the activation computed by each dendritic segment given the context vector corresponding to each task, before (top row) and after (bottom row) training. Note that the dendritic segments for a particular neuron are completely separate of the segments of another in both multi-task RL and continual learning scenarios (e.g., Neuron A's first segment is unrelated to Neuron B's first segment).
}
\label{fig:dendrite_behavior}
\end{figure}

What is the effect of dendrites on a single neuron? In Figure \ref{fig:dendrite_behavior}, we analyze a few Active Dendrites Neurons and their responses to different context vectors before and after learning 10 multi-task RL and permutedMNIST tasks in sequence. At the beginning of training, the responses are random with scattered positive, negative, and near-zero responses. After training, most responses are weak and only a few are either strongly positive or negative. Notably, across the neurons, dendrites only have strong responses to a few contexts as different neurons participate in different subnetworks. We note that in the multi-task RL scenario, we observe both strong positive and negative responses while the continual learning scenario only shows strong positive activity. We are unclear as to why this particular behavior emerges in continual learning but not multi-task RL.

\subsubsection{Impact of Sparsity Level and the Number of Dendrites}
We show that an Active Dendrites Network is competitive with benchmarks in both multi-task RL and continual learning. However, to what extent are active dendrites and sparse representations both contributing factors towards alleviating catastrophic forgetting?

\begin{figure}[h]
\begin{center}
\begin{minipage}{.49\textwidth}
  \centering
  \includegraphics[width=3.0in]{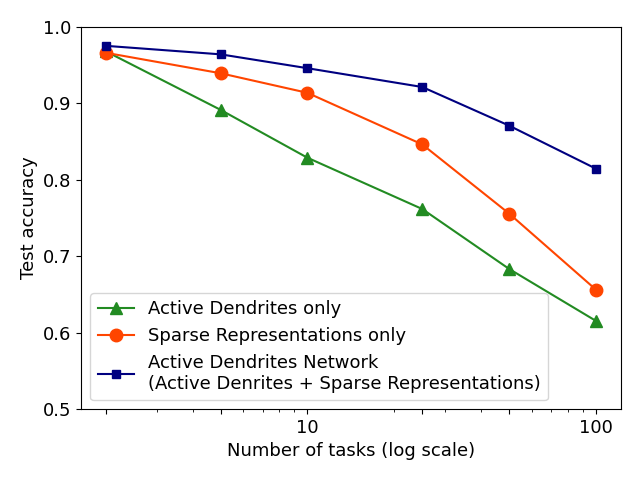}
\end{minipage}
\begin{minipage}{.49\textwidth}
  \centering
  \includegraphics[width=3.0in]{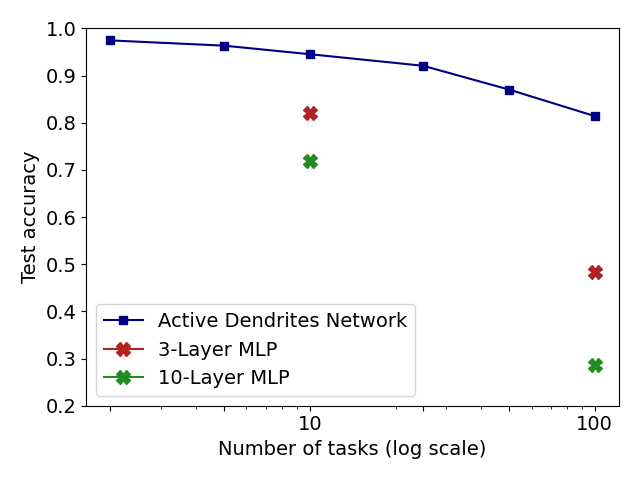}
\end{minipage}
\end{center}
\caption{
{\bf Left:} Continual learning test accuracy on permutedMNIST using active dendrites and dense representations (green), regular ANNs with sparse representations (orange), and Active Dendrites Networks (blue) which use both active dendrites and sparse representations.
{\bf Right:} Continual learning test accuracy for our Active Dendrites Network compared to regular feedforward networks with more layers.
Our Active Dendrites Network has 3 layers; the 2 hidden layers contain neurons modulated by dendritic segments.
In all experiments (left and right subfigures), we average results over 8 independent runs, each with a randomly initialized seed, and omit standard error bars as they highlight a very small range.
}
\label{fig:dendrites_vs_sparsity_and_dendrites_vs_more_layers}
\end{figure}

We investigate this question in the context of continual learning.
We find that both active dendrites without sparse representations and standard point neurons with sparse representations are better than chance in a continual learning scenario.
However, the combination of both active dendrites and sparse representations yield significantly better results than either one on its own.
As Figure~\ref{fig:dendrites_vs_sparsity_and_dendrites_vs_more_layers} (left) shows, the accuracy of both methods evaluated independently and evaluated together on 10 and 100 permutedMNIST tasks demonstrates the importance of implementing both active dendrites and sparse representations.

\begin{figure}[h]
\centering
\includegraphics[width=5.0in]{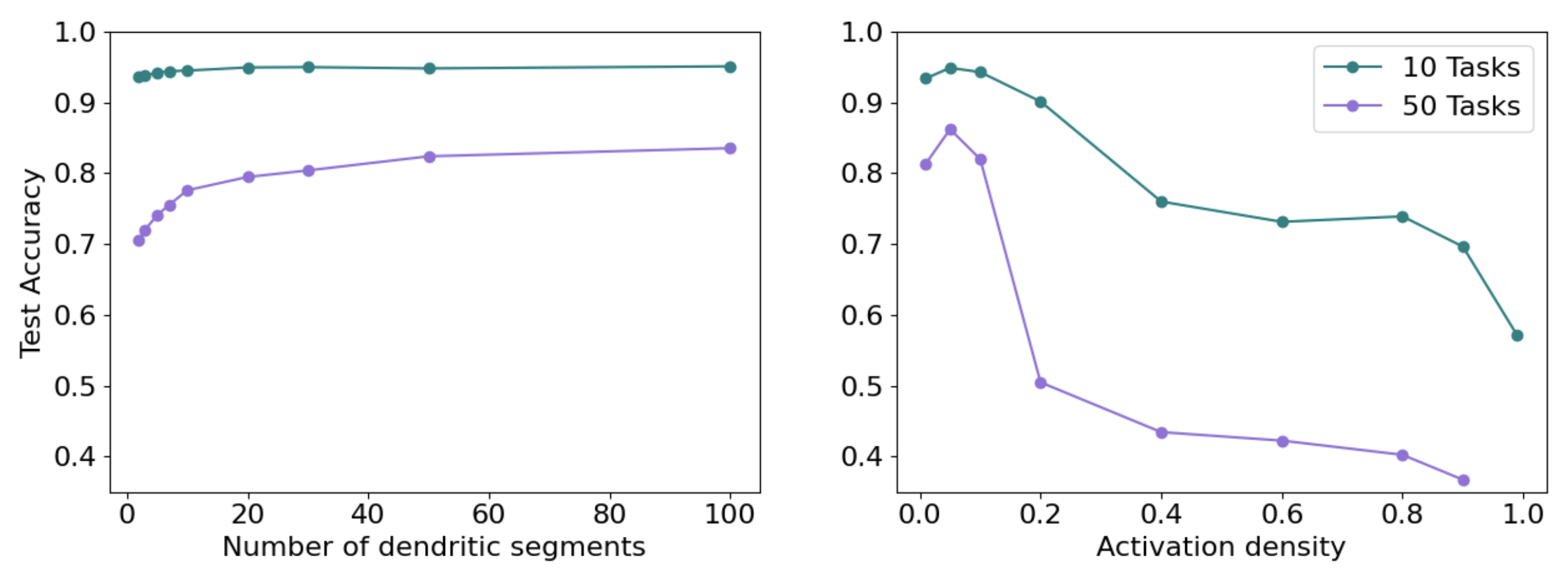}
\caption{
{\bf Left column:} Final accuracy on test examples across all tasks when varying the number of dendritic segments per neuron and keeping activation sparsity constant when learning 10 (top) and 50 (bottom) permutedMNIST tasks.
{\bf Right column:} Final accuracy on test examples across all tasks for a fixed number of dendritic segments per neuron and varying activation density level on 10 (top) and 50 (bottom) permutedMNIST tasks.
}
\label{fig:accuracy_vs_num_segments_sparsity}
\end{figure}

To further test the impact of dendrites and sparsity, we run two additional tests in the continual learning scenario.
First, we fix the level of sparsity in our hidden representations and vary the number of dendritic segments per hidden neuron. Second, we fix the number of dendritic segments per hidden neuron and vary the sparsity in our hidden representations (i.e., vary $k$ in $k$WTA).
As seen in Figure~\ref{fig:accuracy_vs_num_segments_sparsity} (left), increasing the number of dendritic segments leads to a small monotonic increase in accuracy. Figure~\ref{fig:accuracy_vs_num_segments_sparsity} (right) shows that reducing sparsity translates to a sharp drop in accuracy, further highlighting the need for sparse representations.

\subsubsection{Are Networks with Dendrites Equivalent to Larger Networks?}
\label{section:dendrites_vs_more_layers}

Over the last couple of decades, multiple studies have suggested that dendritic computations performed by pyramidal neurons can be approximated by ANNs that have one or more hidden layers.
For example, \citet{poirazi2003} shows that a larger 2-layer neural network can well approximate the post-synaptic responses of a pyramidal neuron with active dendrites. Various follow-up studies also make similar claims \citep{jadi2014, beniaguev2021}, with \citet{beniaguev2021} suggesting that a pyramidal neuron is equivalent to a larger ANN with $7$ hidden layers. In this section we show that in the dynamic scenarios considered here, an Active Dendrites Network is not equivalent to larger or deeper ANNs.

In the case of multi-task RL, a pyramidal neuron's activity \emph{cannot} be approximated by a neural network with more parameters. For instance, classical deep networks that are trained on a variety of tasks are incapable of performing well due to gradient interference, an issue that cannot be solved with simply more hidden neurons. When comparing a 3-layer Active Dendrites Network and a 3-layer MLP with 500,000 more learnable, non-zero parameters, Figure~\ref{fig:mtrl_accuracy} shows that networks with dendrites and sparse representations far outperform the MLP baseline. We also experiment with larger 3-layer MLPs that have 1,700,000 more non-zero parameters than our Active Dendrites Network (hyperparameters found in Table~\ref{table:mtrl_hyperparameters} of Section~\ref{methods:RL_experiment_settings}). In this case, we find that the MLP produces a success rate of \bacc{73.1} across 10 tasks (averaged over the last 500,000 environment steps of training), which underperforms our Active Dendrites Network yielding an average success rate of \bacc{87.5}.

In addition, in the continual learning setting, our network with dendrites cannot be approximated by a neural network with multiple layers. When considering continual learning, classical deep networks are incapable of performing well due to catastrophic forgetting, regardless of network depth. This specific trend can be observed in Figure~\ref{fig:dendrites_vs_sparsity_and_dendrites_vs_more_layers} where our Active Dendrites Network outperforms standard MLPs that have a) the same number of layers but no dendrites (for 10 and 100 permutedMNIST tasks), and b) many more layers and roughly the same number of learnable parameters (for 10 permutedMNIST tasks). (Other ablation studies, not shown in Figure~\ref{fig:dendrites_vs_sparsity_and_dendrites_vs_more_layers}, are described in the Appendix.)

These results for both multi-task RL and continual learning suggest that standard ANNs that are wider or deeper are still prone to gradient interference and catastrophic forgetting while active dendrites can help retain knowledge from previous tasks. In these dynamic settings, our experiments show that a standard feedforward network with more hidden units or additional layers is not as powerful as a network with active dendrites.

\section{Discussion}
\label{section:discussion}

The exact mechanistic details of how a biological neuron converts incoming signals into action potentials (i.e., spikes) remain unclear. Ever since Rosenblatt \citep{rosenblatt1958}, models of biological neurons favor a single linear weighted sum (the point neuron) as a tractable abstraction. This idea continues to serve as the prevalent paradigm in machine learning today for the individual computational unit. One shortcoming is that standard ANNs with point neurons can suffer from catastrophic forgetting. They overwrite many of their connections for each learning iteration, and thus quickly lose previously acquired knowledge \citep{french1999, parisi_continual_2019}. 

In this article we show that augmenting point neurons with biological properties such as active dendrites and sparse representations significantly improves a network's ability to learn multiple tasks at once.  In the multi-task RL setting, a 3-layer Active Dendrites Network can achieve an average accuracy of about \bacc{88} when learning 10 Meta-World tasks together. In the continual learning setting, an almost identical network can achieve greater than \bacc{90} accuracy when learning 100 permutedMNIST tasks in sequence. These results, on two very different scenarios, suggest that Active Dendrites Networks may represent a general purpose architecture for avoiding interference and forgetting in complex settings. In the rest of this Discussion we elaborate on this idea and describe some relationships to other research.

\subsection{Dendrites Enable Dynamic Context Integration and Routing}

\begin{figure}[t]
\centering
\includegraphics[width=5.0in]{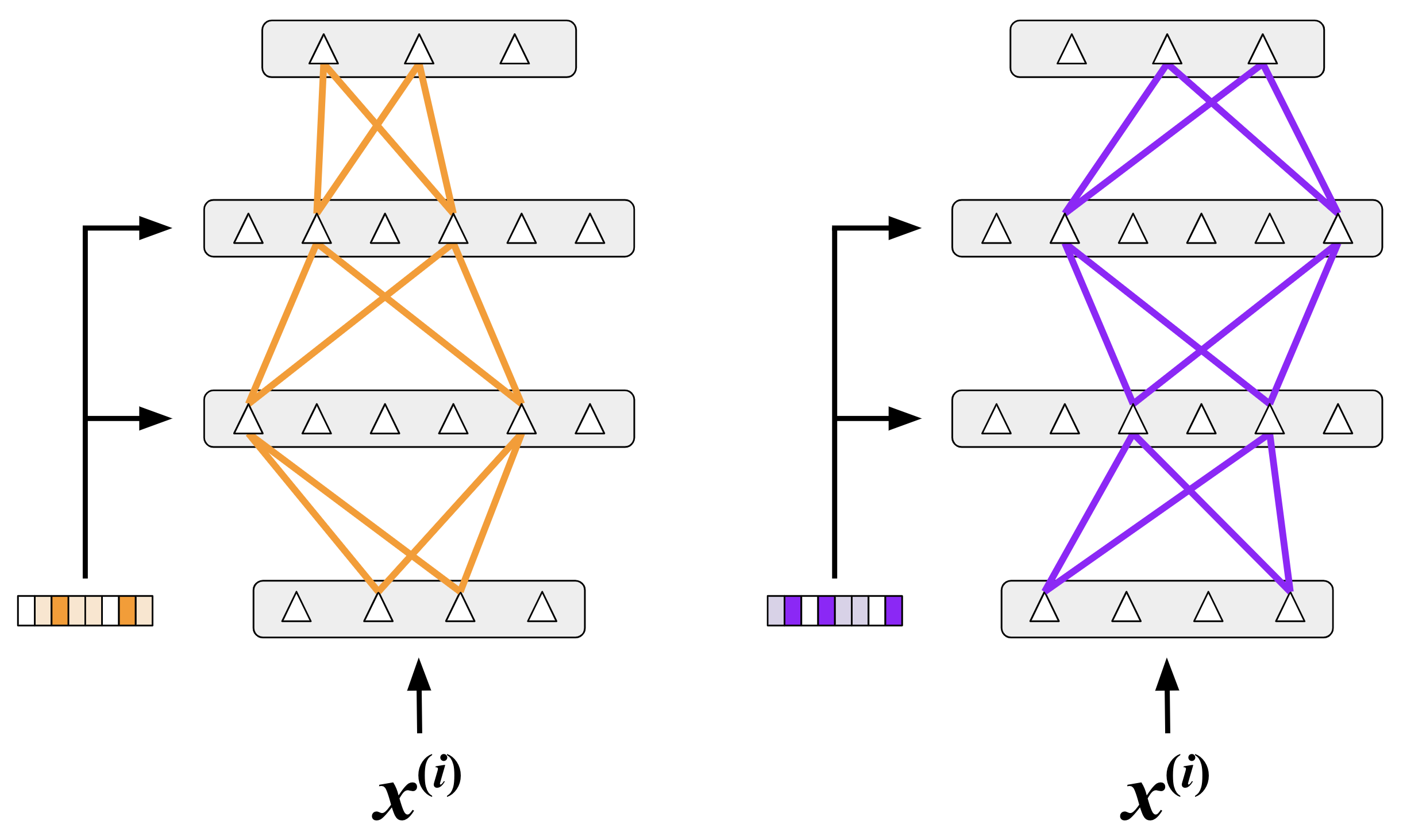}
\caption{
A representation of subnetworks within an Active Dendrites Network.
By receiving different context vectors as input, dendritic segments invoke different subnetworks for a fixed feedforward input. The subnetworks are distributed, i.e., they may share some of the same neurons.
}
\label{fig:subnetwork}
\end{figure}

In this section we attempt to elucidate how active dendrites help in dynamic scenarios such as multi-task and continual learning, and discuss our theory of their underlying role in the neocortex.
Following experimental evidence (Section~\ref{section:active_dendrites}), our model suggests that dendritic segments in each neuron identify specific contexts and then modulate neuronal activity based on this identification.
Combined with subsequent local inhibition ($k$WTA function), the modulation can impact whether the neuron activates.

We propose that the consequence of this behavior is to invoke sparse context-specific subsets of the network. Two different context vectors can lead to different winners and different sparse activation patterns (illustrated in Figure \ref{fig:subnetwork}).
As suggested by the figure, the same feedforward input can activate completely different neurons based on the specific context. Note that the subnetworks are distributed and that two different subnetworks may share some neurons. In Section~\ref{section:subnetwork-results}, we showed that task-specific representations do indeed emerge in our experiments (Figure~\ref{fig:activation_by_task}). 

Why do subnetworks help? In dynamic conditions, the system must react and learn in constantly changing situations. Subnetworks restrict the flow of information to be highly context-dependent and relevant to each specific situation. In addition, errors will only propagate through the active subnetwork. Only the active neurons will update their feedforward weights and only the winning segment within those active neurons will update their dendritic weights. Thus, by utilizing context the brain can isolate information flow, and direct learning itself in a highly localized and task specific manner. The last two decades have seen significant experimental support for highly localized task specific learning in the dendrites of pyramidal neurons \citep{losonczy2008, yang2014, Kerlin2019, limbacher_emergence_2020}.

% In our experiments we show that individual dendritic segments within a neuron learn to respond strongly to different context vectors (Figure~\ref{fig:dendrite_behavior}).  

What is the role of context? In this article we have used a context vector that represents the current task. Prior experimental and modeling work shows the utility of various other types of context. In recurrent networks, it is possible to use the previous activity of the network as context for dendrites. In this case a layer of neurons becomes a powerful sequence memory system \citep{hawkins2016}. For sensorimotor inference, if the coordinates of an external reference frame is used as context, neurons can perform object recognition with actively moving sensors and by integrating information over time \citep{Hawkins2017a}. \citet{Schmidt-Hieber2017} and \citet{Heald2021} also provide experimental evidence for the role of dendrites in separating out information in continuous sensorimotor streams. In the neocortex, if the inference results of neighboring cortical areas are used as context, dendrites can be used to disambiguate uncertain information and perform voting \citep{Hawkins2017a}.

In each of the above scenarios, although the nature of the context greatly impacts emerging behavior, the fundamental operations remain the same. Dendrites recognize patterns that best match their synapses, and up-modulate their neurons such that they are more likely to win. This in turn invokes context specific subnetworks that route information flow and gate learning in order to effectively learn and perform the task at hand. Contextual routing mediated by dendrites may thus be a general-purpose and powerful capability that underlies much of cognitive function \citep{Phillips2015, Phillips2015a}. Indeed, the ability to generate context-dependent output based on a common set of operations could be a crucial building block of cognitive maps able to cover any domain \citep{Whittington2022}. \cite{Flesch2022} provide experimental evidence for contextual gating in a study of human continual learning and memory.

In our implementation we have focused primarily on feed-forward information flow and basal dendrites, and have ignored recurrent and feedback connections and apical dendrites \citep{larkum_new_1999, LARKUM2022}. Interestingly, lateral connections and feedback connections seem to segregate onto different dendritic integration zones \citep{LAFOURCADE2022, Guest2021}. Recent experimental evidence suggests that apical dendrites also process feedback context and have a modulatory impact on the cell leading to task specific functionality \citep{takahashi2020, Kerlin2019, Schoenfeld2021}. From a modeling perspective, there is additional complexity related to generating top-down context \citep{Siegel2000} and simultaneously processing three separate input streams \citep{Phillips2015, LARKUM2022}, an interesting area for future research.

In this article we have focused on modeling the dendritic properties of pyramidal neurons, but we note that dendritic modulation and gating may occur with other neuron types. For example, thalamocortical neurons may exhibit analogous dendrite initiated gating properties \cite{Errington2011}. As such, dendrite mediated contextual integration and gating may be a more general phenomenon of biological neural systems. Modeling other neuron types is an interesting area for future work.

\subsection{Comparison to Other Multi-Task RL Systems}

Many techniques in multi-task RL make manual changes to the network structure or learning scheme in order to account for the learning of new tasks. 
In multi-task scenarios, optimizers struggle to learn different tasks that vary in gradient magnitude and have conflicting gradient direction. In these cases, tasks with larger magnitudes are usually preferred during optimization over others. To rectify this issue, \citet{yu2020gradient} minimizes gradient interference by orthogonally projecting the gradients of tasks that conflict with each other. 
Additionally, in most scenarios, a policy trained on a specific task with a specific agent cannot be adapted to similar problem settings. \citet{devin2017learning} proposes a framework to learn separate policy modules corresponding to a particular task or robotic agent. Ultimately, they show how these modules can be mixed to perform new task--agent combinations or serve as a starting point for good initializations when learning complex behaviors. 
Many multi-task problems also highlight the issue of parameter sharing between distinct tasks. To that end, \citet{yang2020multi} introduces a base policy network composed of multiple modules and a separate routing network. The routing network uses a task embedding and the current state of the agent to reconfigure the base network's modules with a learned routing strategy. 

In contrast, our Active Dendrites Network activates sparse subnetworks by introducing control over individual neurons in a network. By dynamically integrating a context vector to modulate these neurons, the network automatically creates distinct subnetworks to learn each task. Unlike prior approaches, our network does not require modified learning rules, separate modules, or dedicated routing networks to train new tasks. Rather, a single architecture is capable of reducing gradient interference, learning a diverse range of tasks, and can be applied to scenarios beyond multi-task RL.

\subsection{Comparison to Other Continual Learning Systems}

There are a few papers on continual learning that are very related to the core ideas in this paper. Our networks create representations composed of different sparse subnetworks of neurons. \cite{Abbasi2022} use $k$WTA in conjunction with a modified gradient update method to avoid task interference. XdG \citep{masse2018} and Supermasks \citep{wortsman2020} also explicitly utilize sparse subnetworks per task. XdG, discussed extensively in Section~\ref{section:xdg}, hard-codes a sparse subnetwork for each task. This extra supervision step removes the need to dynamically gate activations but requires knowledge of the task identity during inference. In addition, as seen in Figure~\ref{fig:accuracy}, XdG does not scale as well as our networks. In contrast, Supermasks uses a randomly initialized network and focuses on locating the best subnetwork for each task and forgoes any further training. The technique shows impressive scaling behavior, but it's unclear whether complex tasks can be solved without any network training.

Our Active Dendrites Neurons dynamically determine a representation for each feedforward input based on auxiliary contextual inputs. In the case where the modulation function $f$ involves multiplication, our Active Dendrites Networks are an instance of multiplicative networks. \citet{jayakumar2019} demonstrated that multiplicative networks can excel in multi-task scenarios by learning dynamic representations in a task-specific manner. 

Several ANN-based techniques leverage the idea of auxiliary contextual inputs. For instance, Gated Linear Networks \citep{veness2021} and Dendritic Gated Networks \citep{sezener2021} gate activation values for each neuron based on contextual information. Although inspired by dendrites these models 1) don't activate sparse subnetworks, 2) have fixed random dendritic weights (to model cerebellar dendritic branches), and 3) are binary classifiers (i.e., 10 Dendritic Gated Networks are required to classify MNIST digits). Furthermore, because \citet{sezener2021} test Dendritic Gated Networks only up to 10 permutedMNIST tasks using a very different metric, we cannot provide a direct comparison with our model.

\subsection{Future Work}

Our initial results show that active dendrites and sparse representations can mitigate catastrophic forgetting and interference in multi-task RL and continual learning settings. One crucial next step is to test this framework on more real-world scenarios with greater complexity than MT10 or permutedMNIST. The majority of existing work in MTRL considers tasks with shared input and action spaces. Dendrites may be beneficial in scenarios where this assumption does not hold. Extending to tasks with very different input and output spaces is an interesting area for future research. Another interesting area is to combine our two scenarios and explore continual multi-task RL. While testing on more diverse benchmarks, it will also be important to explore additional methods for generating context vectors for a given task. Another important direction for future research is to investigate sparse dendritic segments, following neuroscience evidence suggesting that each segment relies on just a handful of synapses \citep{Branco2011a}.

\section{Methods}
\label{section:methods}

\subsection{Multi-Task Reinforcement Learning Experiments}
\label{methods:RL}

In this section, we provide the details of our multi-task RL experiments\footnote{PyTorch source code for our experiments is available at \url{https://github.com/numenta/htmpapers}}. We use the Multi-Task Soft Actor-Critic algorithm (MTSAC) originally discussed in \citet{yu2019}, which is described as an adaptation of the Soft Actor-Critic algorithm (SAC) \citep{haarnoja2018soft}. We adapt the code in the original Meta-World GitHub repository\footnote{Meta-World source code is available at \url{https://github.com/rlworkgroup/metaworld}} to fit our experiments.

\subsubsection{Basics of Reinforcement Learning}
\label{methods:basics of RL}

To formalize our specific RL problem, we define some fundamental concepts. The state of the RL agent and the action it will take at a specific time $t$ are denoted as $s_t$ and $a_t$, respectively. The RL algorithm trains a policy $\pi$ to take $a_t$ given $s_t$ in order to maximize total return $G = \sum_t \gamma^t r(a_t, s_t)$ across all time-steps $t$, where $r(a_t, s_t)$ is the reward given by the environment and $\gamma$ is a discount factor to strongly consider immediate rewards.

To optimize this policy, our RL formulation uses Markov Decision Processes (MDPs) to model decision making in stochastic environments. Following the notation introduced in \citet{sutton2018reinforcement}, we consider a finite-horizon MDP defined by the tuple $(S, A, P, r, T)$ that operates in a state space $S$ and action space $A$. The MDP also uses the transition probability $P$ between any two states $s_t$ and $s_{t+1}$ by taking action $a_t$, which is explicitly defined across all states and actions as $P(s_{t+1} | s_t, a_t) : S \times A \rightarrow \mathbb{R}$. Agents in this setting receive a reward $r: S \times A \rightarrow \mathbb{R}$ that is also defined across all states and actions. Additionally, agents must make decisions within a fixed number of steps, denoted by the finite-time horizon $T$.  

The RL algorithm we consider computes a value function that estimates the total return accrued at a specific state. More precisely, the value function describes the significance of starting at some state $s_t$ and following some policy $\pi$. The value function for policy $\pi$ can be defined below:
\begin{align} \label{eq:value function}
    V_\pi (s_t) &= \mathbb{E}_{a_t \sim \pi} \left[ r(s_t, a_t) + \mathbb{E}_{s_{t+1} \sim P} \left[ \gamma V_{\pi}(s_{t+1}) \right] \right] \\
    &= \mathbb{E}_{a_t \sim \pi} \left[ r(s_t, a_t) + \sum_{s_{t+1} \in S} P(s_{t+1} | s_t, a_t) \left( \gamma V_{\pi}(s_{t+1}) \right) \right] \\
    &= \sum_{a_t \in A} \pi (a_t | s_t) \left[ r(s_t, a_t) + \sum_{s_{t+1} \in S}  P(s_{t+1} | s_t, a_t) \left( \gamma V_{\pi}(s_{t+1}) \right) \right]
\end{align}

\noindent Note that Equation~\eqref{eq:value function} establishes a recursive relation with respect to the function $V_\pi$. To estimate the value at a given state $s_t$, an agent must take an action $a_t$ sampled from policy $\pi$ to calculate the expected value at the next state $s_{t+1}$. By repeating this process until a terminal state is reached, the agent can use the value function to choose actions that lead to highly valued states.

The RL algorithm we consider also estimates an action-value function $Q_\pi$. While value functions estimate the value of starting at $s_t$ and following $\pi$, action-value functions estimate the value of starting at $s_t$, taking action $a_t$, and \textit{then} following $\pi$ until a terminal state is reached. This is known as the $Q$ function, which can be described explicitly below: 
\begin{align}
    Q_\pi (s_t, a_t) &= r(s_t, a_t) + \mathbb{E}_{s_{t+1} \sim P} \left[ \gamma \mathbb{E}_{a_{t+1} \sim \pi} \left[ Q_\pi(s_{t+1}, a_{t+1}) \right]  \right]
\end{align}

\noindent Fundamentally, value functions and $Q$ functions can be related by the following two expressions:
\begin{align}
    Q_\pi (s_t, a_t) &= r(s_t, a_t) + \mathbb{E}_{s_{t+1} \sim P} \left[ \gamma V_{\pi}(s_{t+1}) \right] \\
    V_\pi (s_t) &= \mathbb{E}_{a_t \sim \pi} \left[ Q_\pi (s_t, a_t) \right]
\end{align}

\noindent Throughout the training process, explored state, action, reward, and next state transitions--namely ($s_t$, $a_t$, $r_t$, $s_{t+1}$)--are used to train $\pi$. In some algorithms, including ours, these transitions are stored in a replay buffer $\mathcal{D}$ and are sampled by batch during each step of training to dynamically compute either the value or $Q$ function. After a suitable period of exploration, the agent will take actions that yield the maximum $V_\pi (s_t)$ or $Q_\pi (s_t, a_t)$ value. Note that while $V$, $Q$, and $\pi$ are expressed in discrete state and action spaces above, they can be easily extended to work in continuous state and action spaces using function approximations such as neural networks.

\subsubsection{Basics of Multi-Task Reinforcement Learning}

We can extend the ideas in Section~\ref{methods:basics of RL} to our multi-task RL experiments in Meta-World. Specifically, the problem framework uses a separate MDP to model each task $\tau$. In the context of the Meta-World multi-task environment, each task shares identical state and actions spaces and defines common transition probabilities and time horizons. However, each task defines separate reward functions, although all functions share a similar scale and structure to allow a single agent to uniformly learn all tasks. We assume a uniform distribution of tasks $p(\tau)$ and train a task-conditioned, stochastic policy $\pi(a | s, \context)$ to solve all $\mathcal{T}$ tasks, where $\context$ is a context vector that provides information about a specific task. Explicitly, the policy is trained to maximize the total return from the task distribution $p(\tau)$ as expressed by $\mathbb{E}_{\tau \sim p(\tau)} \left[ E_{\pi} \left[ \Sigma_{t=0}^T \gamma^t r_t(s_t, a_t) \right] \right]$. 

\subsubsection{The Multi-Task Soft-Actor Critic Algorithm}
\label{methods:SAC and MTSAC}

The MTSAC algorithm we use in our experiments is based on the SAC algorithm and slightly modified to solve $\tau$ various tasks simultaneously.  In SAC, an RL algorithm uses a $V$ or $Q$ network (known as the critic) to train a policy $\pi$ (known as the actor) to take better actions. SAC modifies the original value function definition to also consider the entropy of the policy $\pi$. By maximizing \textit{both} expected return and entropy, an agent is motivated to explore new states while computing an optimal policy. More details about the SAC algorithm can be found in \citet{haarnoja2018soft}.

In MTSAC, both $\pi$ and $Q$ are conditioned by context vector $\context$ and are thus denoted as $\pi (a_t | s_t, \context)$ and $Q (s_t, a_t | \context)$ respectively. MTSAC also uses $\tau$ different entropy coefficients $\alpha_\tau$ to control the exploration per task. More details about the MTSAC algorithm can be found in \citet{yu2019}.

The Meta-World environment we use is MT10, which contains 10 different tasks that a single robotic arm must solve. All tasks share an identical state space $s_t \in \mathbb{R}^{39}$ and action space $a_t \in \mathbb{R}^4$. Because there are 10 different tasks, $\context$ is a 10-dimensional one-hot encoded vector that describes the task ID.

\subsubsection{Experiment Settings}
\label{methods:RL_experiment_settings}

The training hyperparameters are identical for the Active Dendrites Network and the MLP baseline. For every run, the model is trained for 3,000 epochs. Each epoch comprises of one episode of 500 timesteps for each of the 10 tasks. In total, this amounts to 5,000 timesteps per epoch and 15,000,000 timesteps for the entire run. Our implementation parallels the baseline implementation. In our experiments, we use one Active Dendrites Network to model the policy $\pi$ and another to model the $Q$ function.

The model is also used to collect new data to be stored in a replay buffer. At the end of each epoch, the model is then trained for 250 gradient steps. For each gradient step, the algorithm randomly samples a batch of 2,560 experiences from the replay buffer. The replay buffer is a queue of limited size, capped at 1 million, with newer experiences replacing older experiences.

To allow a better comparison between the models, we set the learning rates, target $Q$ function update rate, and policy minimum and maximum standard deviations to be the same for all runs. The difference between the models is the network architecture of the policy and $Q$ functions. The Active Dendrites Network modulates each neuron in the second hidden layer with 10 dendritic segments, where each segment is a vector of size 10. In total, this dendritic layer adds an additional 280,000 parameters to the overall network. We apply a fixed sparsity mask of 10\% to the weights of the feedforward layers of the Active Dendrites Network to reduce the number of free parameters and keep it of comparable size to the MLP baseline. The Active Dendrites Network also uses a $k$WTA activation function instead of ReLU, which effectively selects the top 25\% of units and zeroes out the remaining during every forward step.

\begin{table*}[h]
\renewcommand*{\arraystretch}{1.3}
\vskip 0.15in

\begin{center}
\begin{tabular}{ l c c c }
\toprule
 & Active Dendrites Network & MLP Baseline & Large MLP\\
\midrule
\multicolumn{3}{l}{Network Hyperparameters}\\
\bottomrule
Feedforward Input Size & 39 & 49 & 49\\
% Context Vector Size & & & \\
Hidden Sizes & 2 $\times$ [2,800] & 2 $\times$ [2,800] & 2 $\times$ [3,000]\\
Output Size & 4 & 4 & 4\\
Feedforward Weight Sparsity & 10\% & 0\% & 0\%\\
Activation Function & $k$WTA & ReLU & ReLU\\
Activation Sparsity & 25\%  & $\sim$ 50\% & $\sim$ 50\%\\
Num. Dendritic Segments per Neuron & 10 & 0 & 0\\
Num. Weights per Dendritic Segment & 10 & --- & --- \\
Num. Hidden Layers Modulated & 1 & --- & ---\\
Dendritic Segment Weight Sparsity & 0\% & --- & --- \\
Non-zero Feedforward Parameters & 7,169,964 & 7,994,004 & 9,165,004\\ 
Non-zero Dendritic Parameters & 280,000 & 0 & 0\\
Non-zero Parameters (total) & 7,449,964 & 7,994,004 & 9,165,004\\

\midrule
\multicolumn{3}{l}{Training Hyperparameters}\\
\bottomrule
Number of Epochs & 3,000 & 3,000 & 3,000\\
Number of Timesteps & 15,000,000 & 15,000,000 & 15,000,000\\
Number of Gradient Steps Per Epoch & 250 & 250 & 250\\
Buffer Sampling Batch Size & 2,560 & 2,560 & 2,560\\
Replay Buffer Size & 1,000,000 & 1,000,000 & 1,000,000\\
Policy Learning Rate & $3 \times 10^{-4}$ & $3 \times 10^{-4}$ & $3 \times 10^{-4}$\\
Q-Function Learning Rate & $3 \times 10^{-4}$ & $3 \times 10^{-4}$ & $3 \times 10^{-4}$\\
Target Q-function Update Rate &  $5 \times 10^{-3}$ & $5 \times 10^{-3}$ & $5 \times 10^{-3}$\\
Policy (Min, Max) Std &  $e^{-20}$, $e^{2}$ &  $e^{-20}$, $e^{2}$ & $e^{-20}$, $e^{2}$\\
Action Sampling Distribution Type &  Tanh Normal & Tanh Normal & Tanh Normal\\
\bottomrule
\end{tabular}
\end{center}
\vskip -0.1in
\caption{The hyperparameters for each multi-task RL model.}
\label{table:mtrl_hyperparameters}
\end{table*}

\subsection{Continual Learning Experiments}
\label{methods:continual learning}

% STRUCTURE OF THIS SECTION
% * Specifics of our continual learning scenario
% * Details about permutedMNIST
% * Hyperparameters

We discuss the setup of the continual learning experiments.
Our model is trained on $\mathcal{T}$ discrete tasks in sequence.
More specifically, our model first trains on task $\tau = 1$.
Once learning task $\tau$ is complete, the model then starts training on task $\tau + 1$.
After training on task $\tau = \mathcal{T}$, all learning is complete.
Each task $\tau$ consists of standard batch learning with i.i.d. training data.
While training on task $\tau$ where $1 \leq \tau \leq \mathcal{T}$, our model only receives training data corresponding to task $\tau$.
Once the model finishes learning task $\tau$, it never again receives information about any task $\tau' \leq \tau$ for training purposes.
The model is, however, evaluated on the test data for each task to determine how well it performs.

\subsubsection{permutedMNIST}

We train our model on the permutedMNIST dataset, a benchmark dataset for continual learning \citep{goodfellow2014}, which is derived from MNIST.
MNIST comprises approximately 60,000 black and white images of handwritten digits 0--9 where each such image has dimensions $28 \times 28$ pixels and the associated target digit as the label.
During training, roughly 50,000 images are used for training and the remaining 10,000 for testing.

\begin{figure}[h]
\centering
\includegraphics[width=3.6in]{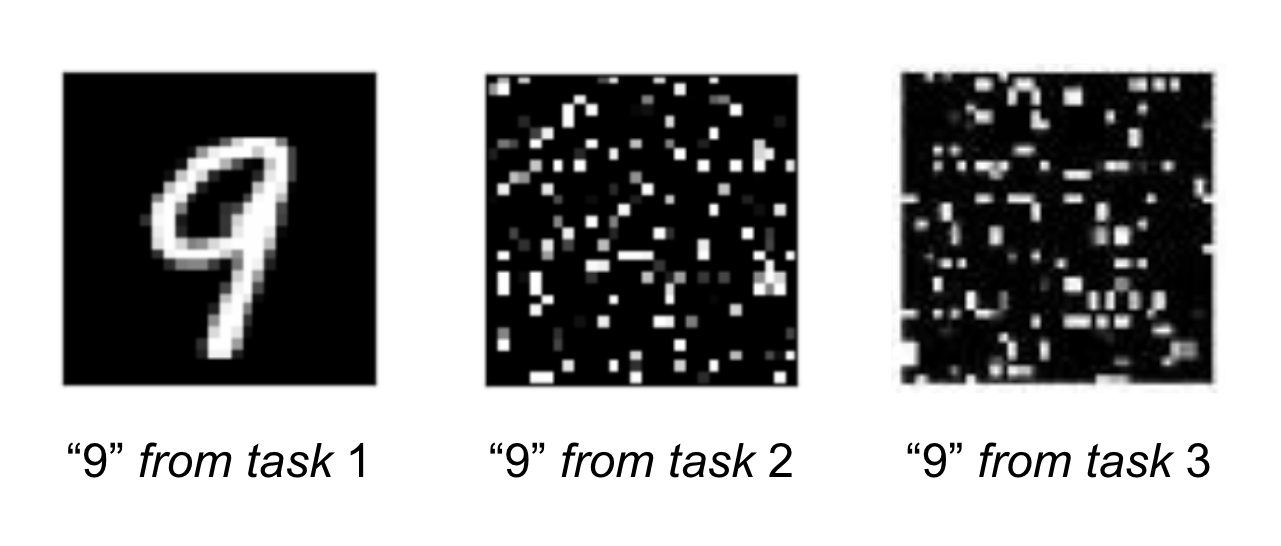}
\caption{
A visual illustration of permutedMNIST.
Each task applies a unique pixel-wise permutation to the same original image (leftmost image) while preserving the target label.
A model's task is to identify the digit in each case regardless of permutation.
}
\label{fig:permutedmnist}
\end{figure}

In permutedMNIST with $\mathcal{T}$ tasks, MNIST is replicated $\mathcal{T}$ times, but each time with a unique pixel-wise permutation applied to all 60,000 images.
That is, each task randomly re-arranges the pixels of all images exactly the same way while preserving the associated target label.
The first task ($\tau = 1$) corresponds to the identity permutation (i.e., regular MNIST) and every subsequent task generates a random pixel-wise permutation.
As permutedMNIST is synthesized from regular MNIST, there can be an arbitrary number of tasks, $\mathcal{T}$.
Figure \ref{fig:permutedmnist} illustrates a single image taken from different tasks.

Our model, and all comparisons we made, uses a single output head.
Each model has 10 output units in the final layer of the network representing the 10 categories. These output units are re-used for each task, i.e., the model is trained to predict the first output unit for label ``0'' regardless of which task the input data corresponds to. In this setup chance accuracy is $10\%$.

\subsubsection{Experiment Settings}

When employing the prototype method described in Section \ref{section:context} to select context signals at test time only, we train an Active Dendrites Network with 2 hidden layers that comprise Active Dendrites Neurons.
We find that having just a single hidden layer reduced accuracy by a few percentage points while 3 hidden layers provided a minimal performance boost. For 100 tasks, a single layer reduced accuracy by $3\%$ and three layers improved accuracy by $0.5\%$.
For all training, we use the Adam optimizer \citep{kingma2015} and a batch size of 256 samples.
Table \ref{table:cl_hyperparameters} gives the exact hyperparameters and model architecture for each model we train and evaluate on permutedMNIST. Note that hyperparameters were optimized individually for each setting.

\begin{table*}[h]
\renewcommand*{\arraystretch}{1.3}
\vskip 0.15in

\begin{center}
\begin{small}
\begin{tabular}{l|P{26mm}P{26mm}P{20mm}P{21mm}}
\toprule
& Active Dendrites Network {\footnotesize (task info. provided)} & Active Dendrites Network {\footnotesize (task info. not provided)} & 3-Layer MLP & 10-Layer MLP\\
\midrule
\multicolumn{3}{l}{Network Hyperparameters}\\
\bottomrule
Feedforward Input Size & 784 & 784 & 784 & 784\\
% Context Vector Size & 784 & 256 & --- & --- \\
Hidden Sizes & 2 $\times$ [2,048] & 2 $\times$ [2,048] & 2 $\times$ [2,048] & 10 $\times$ [2,048]\\
Output Size & 10 & 10 & 10 & 10\\
Feedforward Weight Sparsity & 50\% & 50\% & 0\% & 0\%\\
Activation Function & $k$WTA & $k$WTA & ReLU & ReLU\\
Activation Sparsity & 5\% & 5\% & $\sim$ 50\% & $\sim$ 50\%\\
Num. Dendritic Segments per Neuron & $\mathcal{T}$ & $\mathcal{T}$ & 0 & 0\\
Num. Weights per Dendritic Segment & 784 & 784 & --- & --- \\
Dendritic Segment Weight Sparsity & 0\% & 0\% & --- & --- \\
Non-zero Feedforward Parameters & 2,914,314 & 2,914,314 & 5,824,522 & 35,198,986\\
Non-zero Dendritic Parameters & $\mathcal{T} \times 3,211,264$ & $\mathcal{T} \times 3,211,264$ & 0 & 0\\
Non-zero Parameters (total) & \multicolumn{2}{c}{See Appendix} & 5,824,522 & 35,198,986\\
\midrule
\multicolumn{3}{l}{Training Hyperparameters}\\
\bottomrule
($\mathcal{T}=2$) Learning Rate & $5 \times 10^{-4}$ & $10^{-3}$ & --- & ---\\
($\mathcal{T}=2$) Number of Epochs & 1 & 5 & --- & ---\\
($\mathcal{T}=5$) Learning Rate & $5 \times 10^{-4}$ & $10^{-3}$ & --- & ---\\
($\mathcal{T}=5$) Number of Epochs & 1 & 5 & --- & ---\\
($\mathcal{T}=10$) Learning Rate & $5 \times 10^{-4}$ & $10^{-3}$ & $3 \times 10^{-6}$ & $3 \times 10^{-6}$\\
($\mathcal{T}=10$) Number of Epochs & 3 & 3 & 5 & 3\\
($\mathcal{T}=25$) Learning Rate & $3 \times 10^{-4}$ & $3 \times 10^{-4}$ & --- & ---\\
($\mathcal{T}=25$) Number of Epochs & 5 & 1 & --- & ---\\
($\mathcal{T}=50$) Learning Rate & $3 \times 10^{-4}$ & $10^{-4}$ & --- & ---\\
($\mathcal{T}=50$) Number of Epochs & 3 & 3 & --- & ---\\
($\mathcal{T}=100$) Learning Rate & $10^{-4}$ & $10^{-4}$ & $10^{-6}$ & $3 \times 10^{-7}$\\
($\mathcal{T}=100$) Number of Epochs & 3 & 3 & 3 & 3\\
\bottomrule
\end{tabular}
\end{small}
\end{center}
\vskip -0.1in
\caption{The hyperparameters used to train each model on permutedMNIST.}
\label{table:cl_hyperparameters}
\end{table*}

To combine Active Dendrites Network with SI, and to compare against XdG, we reduce the number of units in each hidden layer from 2,048 to 2,000 as to exactly match the architectures (with the exception of dendritic segments) used in the SI and XdG papers. (See Appendix for a discussion on the number of parameters.)
In addition, the SI-and-Active-Dendrites network is trained for 20 epochs per task instead of just 3 as this significantly improves results.
We fix the learning rate to be $5 \times 10^{-4}$ for all numbers of tasks, and we use SI regularization strength $c = 0.1$ and damping coefficient $\xi = 0.1$.
Both a) training for 20 epochs per task and b) the $c, \xi$ values that we use here align with the training setups of \citet{zenke2017} and \citet{masse2018}.

\subsubsection{Constructing Prototypes During Training Without Task Information}
\label{section:training_prototype}

When task information is not given during training nor testing, the task corresponding to each input example must be inferred. This section describes the online clustering method we implemented to infer task information during training. One inductive bias in our procedure is that all training examples in a batch correspond to the same task, since continual learning scenarios usually only observe examples from a single task within a given batch.

Formally, let $X = \{\x^{(1)}, \ldots, \x^{(n)}\}$ be a batch of $n$ training examples (in the case of permutedMNIST, each $\x^{(i)}$ is a 784-dimensional vector for $1 \leq i \leq n$).
Suppose $M$ individual prototypes are designated thus far: $\p_1, \ldots, \p_M$.
For each $\p_j$ (where $1 \leq j \leq M$), the individual examples used to construct that prototype are also stored in memory: $Y_j = \{\y^{(1)}, \ldots, \y^{(m_j)}\}$, where $m_j$ gives the number of examples for cluster $j$.
These previous training examples are observed by the learner during previous batches of learning and stored in memory.
We identify if the new batch $X$ is similar enough to any cluster of training examples $Y_j$ such that the corresponding prototype $\p_j$ should be used as the context signal.
If a cluster $j$ is found such that $X$ is ``similar'' to $Y_j$, then $Y_j$ is expanded to include $X$.
Subsequently, $\p_j$ is updated to incorporate samples from $X$.
Otherwise, if $X$ is deemed significantly different from $Y_j$ for all $j$, then a new cluster is formed: $Y_{M+1} \gets X$ and its prototype is is the element-wise mean of all $\x \in X$.
Algorithm \ref{alg:cluster} describes the procedure for clustering during training when task information is not provided.

\begin{algorithm}
\caption{
Clustering algorithm by which a new batch of inputs $X$ either gets assigned to 1 of $M$ existing clusters or initiates cluster $M + 1$.
This procedure is greedy since it assigns $X$ to the first cluster $j$ that it suitably matches.
}
\label{alg:cluster}
\begin{algorithmic}[1]

\Procedure{Cluster}{$X$, $Y$}
  \State $M \gets 0$ \Comment{Number of existing clusters}
  
  \While{not done learning}
    \State $X \gets$ new batch
    \State assigned $\gets$ {\bf False}
    
    \For{$j=1$ to $M$}

      \If{$\neg$ assigned and \textsc{Is\_Match}($X$, $Y_j$)}
        \State assigned $\gets$ {\bf True}
        \State $Y_j \gets Y_j \cup X$
        \State update $\p_j$ to include each $\x \in X$
      \EndIf
    \EndFor
  \EndWhile
\EndProcedure
\end{algorithmic}
\end{algorithm}

In the pseudocode, how do we determine when $X$ is similar enough to some $Y_j$?
If we have univariate data (i.e., if each $\x \in X$ and $\y \in Y_j$ is a scalar quantity), we could use an unpaired $t$-test do this.
Instead, we use a generalized version of an unpaired $t$-test that applies to multivariate data. In our hypothesis testing setup, the null hypothesis is that for any given $j$, the same underlying process generates samples from both $X$ and $Y_j$. When we accept the null hypothesis, we assume each $\x \in X$ and each $\y \in Y_j$ are training examples from the same permutedMNIST task---and therefore $\p_j$ can be used as the context signal when training an Active Dendrites Network on examples in $X$ (albeit $\p_j$ is first updated to account for $X$).

\citet{hotelling1931} proposed Hotelling's $t$-squared statistic ($t^2$) as a generalization of the $t$-statistic used to perform single-variable $t$-tests; it is computed as $$t^2 = \frac{|X||Y_j|}{|X| + |Y_j|} \left( \bar{\x} - \bar{\y} \right)^\top \boldsymbol{\Sigma}^{-1} \left( \bar{\x} - \bar{\y} \right)$$ where $\bar{x}$ and $\bar{y}$ are simply the element-wise means of all $\x \in X$ and $\y \in Y_j$, respectively, and $\boldsymbol{\Sigma}$ is the pooled, sample-adjusted covariance matrix of samples in $X$ and $Y_j$.
The test statistic $t^2$ can be compared to a chosen $p$-value to accept or reject the null hypothesis by first transforming it to a value drawn from an $F$-distribution (whose cumulative density function is more well-studied than that of the $t$-squared distribution) as follows: $$f = \frac{|X| + |Y_j| - d - 1}{d \left( |X| + |Y_j| - 2 \right)} t^2$$ where $d$ is the dimensionality of the samples.

We fix a $p$-value and derive a value for $f$ based on $t^2$ as give above.
If $f > p$, then we reject the null hypothesis since the probability that the same generative process explains both $X$ and $Y_j$ is extremely low, and thus create a new cluster.
Since we perform pairwise multivariate $t$-tests between $X$ and $Y_j$ for all existing prototypes $j$, a new cluster and prototype emerge if and only if we reject the null hypothesis for all $M$ $t$-tests.
Algorithm \ref{alg:hotelling} describes the procedure for performing the multivariate $t$-test via the $t$-squared statistic given two sets of multivariate samples.

\begin{algorithm}
\caption{
Unpaired multivariate $t$-test using Hotelling's $t$-squared statistic.
Here, we use a slight abuse of notation when computing covariance matrices by assuming sets of $d$-dimensional vectors can also be treated as matrices whose rows correspond to their $d$-dimensional elements.
We assume a $p$-value is fixed a priori. In our implementation, we replace all standard matrix inversions with the Moore-Penrose pseudo-inversion.
}
\label{alg:hotelling}
\begin{algorithmic}[1]

\Procedure{Is\_Match}{$X$, $Y$}
  \State $\bar{\x} \gets \frac{1}{|X|} \sum_{\x \in X} \x$ \Comment{Compute $X$ mean}
  \State $\bar{\y} \gets \frac{1}{|Y|} \sum_{\y \in Y} \y$ \Comment{Compute $Y$ mean}

  \State $\boldsymbol{\Sigma}_X \gets \frac{1}{|X| - 1} (X - \bar{\x}) (X - \bar{\x})^\top$ \Comment{Compute $X$ covariance}
  \State $\boldsymbol{\Sigma}_Y \gets \frac{1}{|Y| - 1} (X - \bar{\y}) (Y - \bar{\y})^\top$ \Comment{Compute $Y$ covariance}
  
  \State $\boldsymbol{\Sigma} \gets \frac{(|X| - 1) \boldsymbol{\Sigma}_X + (|Y| - 1) \boldsymbol{\Sigma}_Y}{|X| + |Y| - 2}$ \Comment{Compute pooled covariance}
  \State $t^2 \gets \frac{|X||Y_j|}{|X| + |Y_j|} \left( \bar{\x} - \bar{\y} \right)^\top \boldsymbol{\Sigma}^{-1} \left( \bar{\x} - \bar{\y} \right)$ \Comment{Compute $t^2$}
  
  \State $f = \frac{|X| + |Y_j| - d - 1}{d \left( |X| + |Y_j| - 2 \right)} t^2$ \Comment{Convert $t^2$ to $f$}

  \If{$f > p$}:
    \State \Return {\bf False} \Comment{Reject null hypothesis}
  \Else
    \State \Return {\bf True}  \Comment{Accept null hypothesis}
  \EndIf
\EndProcedure

\end{algorithmic}
\end{algorithm}

\subsection{Absolute Max Gating}
\label{section:absmax}

We outline how we implement gating in Active Dendrites Networks. In Section~\ref{section:model}, we present gating as modifying the value of the weighted linear sum computed by the point neuron based on the maximum activation, i.e., $\sigma(\max_j \boldsymbol{u}^\top \boldsymbol{c})$.
One problem with this formulation is that it becomes difficult to turn a neuron off (i.e., force it's activation value to be zero) due to the $\max$ operator.
That is, if dendritic segment $j$ learns to turn off the unit, then based on sigmoidal gating, we should expect that $\boldsymbol{u}_j^\top \boldsymbol{c}$ is a small number with large absolute value (very negative).
However, it's likely that for some other segment $j'$ ($j \neq j'$), $\boldsymbol{u}_{j'}^\top \boldsymbol{c} > 0 > \boldsymbol{u}_{j}^\top \boldsymbol{c}$ which means that segment $j'$ will be selected by the $\max$ operator instead of segment $j$, hence increasing the chance that the neuron will be selected by the $k$WTA process.

This motivates absolute max gating in which the activation with the largest magnitude is selected and its sign is kept.
More formally, a point neuron augmented with absolute max gating computes its output as
\begin{align*}
    j^* &= \argmax_j \left| \boldsymbol{u}_j^\top \boldsymbol{c} \right|,\\
    \hat{y} &= \left( \boldsymbol{w}^\top \x + b \right) \sigma \left(\boldsymbol{u}_{j^*}^\top \boldsymbol{c} \right).
\end{align*}

\section*{Author Contributions}

SA conceived of the overall theory, and the mapping to neuroscience. KG and SA implemented the core dendrites code. KG and JF implemented the continual learning experiments and their analysis. AV originally conceived of the mapping to multi-task RL and designed the original RL experiments. AI, LS, and AV implemented and ran the multi-task RL experiments. SA, AI, AV, and KG contributed to the paper design and wrote the text.

\section*{Acknowledgements}

We thank Jeff Hawkins, Ben Cohen, Greg Maltz, Luiz Scheinkman, and the rest of the Numenta team for helpful feedback throughout the editing process.

\bibliographystyle{plainnat}
\bibliography{ms}

\begin{thebibliography}{102}
\providecommand{\natexlab}[1]{#1}
\providecommand{\url}[1]{\texttt{#1}}
\expandafter\ifx\csname urlstyle\endcsname\relax
  \providecommand{\doi}[1]{doi: #1}\else
  \providecommand{\doi}{doi: \begingroup \urlstyle{rm}\Url}\fi

\bibitem[Sch(2022)]{Schoenfeld2021}
{Dendritic integration of sensory and reward information facilitates learning}.
\newblock \emph{bioRxiv}, 2022.
\newblock \doi{10.1101/2021.12.28.474360}.
\newblock URL
  \url{https://www.biorxiv.org/content/early/2022/01/12/2021.12.28.474360}.

\bibitem[Abbasi et~al.(2022)Abbasi, Nooralinejad, Braverman, Pirsiavash, and
  Kolouri]{Abbasi2022}
Ali Abbasi, Parsa Nooralinejad, Vladimir Braverman, Hamed Pirsiavash, and
  Soheil Kolouri.
\newblock {Sparsity and Heterogeneous Dropout for Continual Learning in the
  Null Space of Neural Activations}.
\newblock mar 2022.
\newblock \doi{10.48550/arxiv.2203.06514}.
\newblock URL \url{https://arxiv.org/abs/2203.06514}.

\bibitem[Ahmad and Hawkins(2016)]{ahmad2016}
Subutai Ahmad and Jeff Hawkins.
\newblock How do neurons operate on sparse distributed representations? {A}
  mathematical theory of sparsity, neurons and active dendrites.
\newblock \emph{arXiv:1601.00720}, 2016.

\bibitem[Ahmad and Scheinkman(2019)]{ahmad2019}
Subutai Ahmad and Luiz Scheinkman.
\newblock How can we be so dense? {T}he benefits of using highly sparse
  representations.
\newblock \emph{arXiv:1903.11257}, 2019.

\bibitem[Andreas et~al.(2017)Andreas, Klein, and Levine]{andreas2017modular}
Jacob Andreas, Dan Klein, and Sergey Levine.
\newblock Modular multitask reinforcement learning with policy sketches.
\newblock In \emph{Proceedings of the 34th International Conference on Machine
  Learning}, Sydney, Australia, 2017.

\bibitem[Antic et~al.(2010)Antic, Zhou, Moore, Short, and Ikonomu]{antic2010}
Srdjan~D. Antic, Wen-Liang Zhou, Anna~R. Moore, Shaina~M. Short, and
  Katerina~D. Ikonomu.
\newblock The decade of the dendritic {NMDA} spike.
\newblock \emph{Journal of Neuroscience Research}, 88\penalty0 (14):\penalty0
  2991--3001, 2010.
\newblock \doi{10.1002/jnr.22444}.

\bibitem[Antic et~al.(2018)Antic, Hines, and Lytton]{antic2018}
Srdjan~D. Antic, Michael Hines, and William~W. Lytton.
\newblock Embedded ensemble encoding hypothesis: The role of the “prepared”
  cell.
\newblock \emph{Journal of Neuroscience Research}, 96\penalty0 (9):\penalty0
  1543--1559, 2018.
\newblock ISSN 10974547.
\newblock \doi{10.1002/jnr.24240}.

\bibitem[Arulkumaran et~al.(2017)Arulkumaran, Deisenroth, Brundage, and
  Bharath]{arullkumaran2017}
Kai Arulkumaran, Marc~Peter Deisenroth, Miles Brundage, and Anil~Anthony
  Bharath.
\newblock A brief survey of deep reinforcement learning.
\newblock \emph{IEEE Signal Processing Magazine}, 34\penalty0 (6):\penalty0
  26--38, 2017.
\newblock ISSN 1053-5888.
\newblock \doi{10.1109/msp.2017.2743240}.
\newblock URL \url{http://dx.doi.org/10.1109/MSP.2017.2743240}.

\bibitem[Attwell and Laughlin(2001)]{attwell2001}
David Attwell and Simon~B. Laughlin.
\newblock An energy budget for signaling in the grey matter of the brain.
\newblock \emph{Journal of Cerebral Blood Flow and Metabolism}, 21\penalty0
  (10):\penalty0 1133--1145, 2001.
\newblock ISSN 0271678X.
\newblock \doi{10.1097/00004647-200110000-00001}.

\bibitem[Barth and Poulet(2012)]{barth2012}
Alison~L. Barth and James F~a Poulet.
\newblock Experimental evidence for sparse firing in the neocortex.
\newblock \emph{Trends in Neurosciences}, 35\penalty0 (6):\penalty0 345--355,
  2012.
\newblock ISSN 01662236.
\newblock \doi{10.1016/j.tins.2012.03.008}.

\bibitem[Beniaguev et~al.(2021)Beniaguev, Segev, and London]{beniaguev2021}
David Beniaguev, Idan Segev, and Michael London.
\newblock Single cortical neurons as deep artificial neural networks.
\newblock \emph{Neuron}, 109\penalty0 (17):\penalty0 2727--2739, 2021.
\newblock \doi{10.1016/j.neuron.2021.07.002}.

\bibitem[Bentivoglio and Swanson(2001)]{bentivoglio_fine_2001}
Marina Bentivoglio and Larry~W. Swanson.
\newblock On the fine structure of the pes hippocampi major.
\newblock \emph{Brain Research Bulletin}, 54\penalty0 (5):\penalty0 461--483,
  2001.
\newblock ISSN 0361-9230.
\newblock \doi{10.1016/S0361-9230(01)00430-0}.
\newblock URL
  \url{https://www.sciencedirect.com/science/article/pii/S0361923001004300}.

\bibitem[Branco and H{\"{a}}usser(2011)]{Branco2011a}
Tiago Branco and Michael H{\"{a}}usser.
\newblock {Synaptic integration gradients in single cortical pyramidal cell
  dendrites.}
\newblock \emph{Neuron}, 69\penalty0 (5):\penalty0 885--92, 2011.
\newblock ISSN 1097-4199.
\newblock \doi{10.1016/j.neuron.2011.02.006}.

\bibitem[Branco and Häusser(2010)]{branco_single_2010}
Tiago Branco and Michael Häusser.
\newblock The single dendritic branch as a fundamental functional unit in the
  nervous system.
\newblock \emph{Current Opinion in Neurobiology}, 20\penalty0 (4):\penalty0
  494--502, 2010.
\newblock ISSN 0959-4388.
\newblock \doi{10.1016/j.conb.2010.07.009}.
\newblock URL
  \url{https://www.sciencedirect.com/science/article/pii/S0959438810001170}.

\bibitem[Caruana(1997)]{caruana1997}
Rich Caruana.
\newblock Multitask learning.
\newblock \emph{Machine Learning}, 28\penalty0 (1):\penalty0 41--75, 1997.
\newblock \doi{10.1023/A:1007379606734}.
\newblock URL
  \url{http://www.springerlink.com/content/x4q010h7342j4p15/fulltext.pdf}.

\bibitem[Chen et~al.(2018)Chen, Badrinarayanan, Lee, and
  Rabinovich]{chen2018gradnorm}
Zhao Chen, Vijay Badrinarayanan, Chen-Yu Lee, and Andrew Rabinovich.
\newblock Gradnorm: Gradient normalization for adaptive loss balancing in deep
  multitask networks.
\newblock In \emph{Proceedings of the 35th International Conference on Machine
  Learning}, Stockholm, Sweden, 2018.

\bibitem[Cui et~al.(2017)Cui, Ahmad, and Hawkins]{Cui2017}
Yuwei Cui, Subutai Ahmad, and Jeff Hawkins.
\newblock {The HTM Spatial Pooler – a neocortical algorithm for online sparse
  distributed coding}.
\newblock \emph{Frontiers in Computational Neuroscience}, 11:\penalty0 111,
  2017.
\newblock ISSN 1662-5188.
\newblock \doi{10.3389/FNCOM.2017.00111}.

\bibitem[Devin et~al.(2017)Devin, Gupta, Darrell, Abbeel, and
  Levine]{devin2017learning}
Coline Devin, Abhishek Gupta, Trevor Darrell, Pieter Abbeel, and Sergey Levine.
\newblock Learning modular neural network policies for multi-task and
  multi-robot transfer.
\newblock In \emph{Proceedings of the IEEE International Conference on Robotics
  and Automation}, Singapore, 2017.

\bibitem[Dong et~al.(2015)Dong, Wu, He, Yu, and Wang]{dong2015multi}
Daxiang Dong, Hua Wu, Wei He, Dianhai Yu, and Haifeng Wang.
\newblock Multi-task learning for multiple language translation.
\newblock In \emph{Proceedings of the 53rd Annual Meeting of the Association
  for Computational Linguistics and the 7th International Joint Conference on
  Natural Language Processing}, pages 1723--1732, Beijing, China, 2015.
  Association for Computational Linguistics.
\newblock \doi{10.3115/v1/P15-1166}.
\newblock URL \url{https://www.aclweb.org/anthology/P15-1166}.

\bibitem[Du et~al.(2020)Du, Czarnecki, Jayakumar, Farajtabar, Pascanu, and
  Lakshminarayanan]{du2020adapting}
Yunshu Du, Wojciech~M. Czarnecki, Siddhant~M. Jayakumar, Mehrdad Farajtabar,
  Razvan Pascanu, and Balaji Lakshminarayanan.
\newblock Adapting auxiliary losses using gradient similarity.
\newblock \emph{arXiv:1812.02224}, 2020.

\bibitem[Errington and Connelly(2011)]{Errington2011}
A.~C. Errington and W.~M. Connelly.
\newblock {Dendritic T-type Ca2+ Channels: Giving a Boost to Thalamic Reticular
  Neurons}.
\newblock \emph{Journal of Neuroscience}, 31\penalty0 (15):\penalty0
  5551--5553, 2011.
\newblock ISSN 0270-6474.
\newblock \doi{10.1523/JNEUROSCI.0067-11.2011}.

\bibitem[Flesch et~al.(2022)Flesch, Nagy, Saxe, and Summerfield]{Flesch2022}
Timo Flesch, David~G Nagy, Andrew Saxe, and Christopher Summerfield.
\newblock {Modelling continual learning in humans with Hebbian context gating
  and exponentially decaying task signals}, 2022.
\newblock URL \url{https://arxiv.org/abs/2203.11560}.

\bibitem[French(1999)]{french1999}
Robert~M. French.
\newblock Catastrophic forgetting in connectionist networks.
\newblock \emph{Trends in Cognitive Sciences}, 3\penalty0 (4):\penalty0
  128--135, 1999.
\newblock ISSN 1364-6613.
\newblock \doi{10.1016/S1364-6613(99)01294-2}.
\newblock URL
  \url{https://www.sciencedirect.com/science/article/abs/pii/S1364661399012942}.

\bibitem[Gao et~al.(2021)Gao, Graham, Zhou, Jang, Angulo, Dura-Bernal, Hines,
  Lytton, and Antic]{gao2021}
Peng~P. Gao, Joseph~W. Graham, Wen~Liang Zhou, Jinyoung Jang, Sergio Angulo,
  Salvador Dura-Bernal, Michael Hines, William~W. Lytton, and Srdjan~D. Antic.
\newblock Local glutamate-mediated dendritic plateau potentials change the
  state of the cortical pyramidal neuron.
\newblock \emph{Journal of Neurophysiology}, 125\penalty0 (1):\penalty0 23--42,
  2021.
\newblock ISSN 15221598.
\newblock \doi{10.1152/JN.00734.2019}.

\bibitem[Goodfellow et~al.(2014)Goodfellow, Mirza, Xiao, Courville, and
  Bengio]{goodfellow2014}
Ian~J. Goodfellow, Mehdi Mirza, Da~Xiao, Aaron Courville, and Yoshua Bengio.
\newblock An empirical investigation of catastrophic forgetting in
  gradient-based neural networks.
\newblock In \emph{Proceedings of the 2nd International Conference on Learning
  Representations}, Banff, Canada, 2014.

\bibitem[Goyal et~al.(2020)Goyal, Sodhani, Binas, Peng, Levine, and
  Bengio]{goyal2020reinforcement}
Anirudh Goyal, Shagun Sodhani, Jonathan Binas, Xue~Bin Peng, Sergey Levine, and
  Yoshua Bengio.
\newblock Reinforcement learning with competitive ensembles of
  information-constrained primitives.
\newblock In \emph{Proceedings of the 8th International Conference on Learning
  Representations}, Digital, 2020.
\newblock URL \url{https://openreview.net/forum?id=ryxgJTEYDr}.

\bibitem[Guest et~al.(2021)Guest, Bast, Narayanan, and Oberlaender]{Guest2021}
Jason~M Guest, Arco Bast, Rajeevan~T Narayanan, and Marcel Oberlaender.
\newblock {Thalamus gates active dendritic computations in cortex during
  sensory processing}.
\newblock \emph{bioRxiv}, 2021.
\newblock \doi{10.1101/2021.10.21.465325}.
\newblock URL
  \url{https://www.biorxiv.org/content/early/2021/10/21/2021.10.21.465325}.

\bibitem[Haarnoja et~al.(2018{\natexlab{a}})Haarnoja, Pong, Zhou, Dalal,
  Abbeel, and Levine]{haarnoja2018composable}
Tuomas Haarnoja, Vitchyr Pong, Aurick Zhou, Murtaza Dalal, Pieter Abbeel, and
  Sergey Levine.
\newblock Composable deep reinforcement learning for robotic manipulation.
\newblock In \emph{Proceedings of the IEEE International Conference on Robotics
  and Automation}, Brisbane, Australia, 2018{\natexlab{a}}.

\bibitem[Haarnoja et~al.(2018{\natexlab{b}})Haarnoja, Zhou, Abbeel, and
  Levine]{haarnoja2018soft}
Tuomas Haarnoja, Aurick Zhou, Pieter Abbeel, and Sergey Levine.
\newblock Soft actor-critic: Off-policy maximum entropy deep reinforcement
  learning with a stochastic actor.
\newblock In \emph{Proceedings of the 35th International Conference on Machine
  Learning}, Stockholm, Sweden, 2018{\natexlab{b}}.

\bibitem[Hawkins and Ahmad(2016)]{hawkins2016}
Jeff Hawkins and Subutai Ahmad.
\newblock Why neurons have thousands of synapses, a theory of sequence memory
  in neocortex.
\newblock \emph{Frontiers in Neural Circuits}, 10\penalty0 (23):\penalty0
  1--13, 2016.
\newblock \doi{10.3389/fncir.2016.00023}.

\bibitem[Hawkins et~al.(2017)Hawkins, Ahmad, and Cui]{Hawkins2017a}
Jeff Hawkins, Subutai Ahmad, and Yuwei Cui.
\newblock {A Theory of How Columns in the Neocortex Enable Learning the
  Structure of the World}.
\newblock \emph{Frontiers in Neural Circuits}, 11:\penalty0 81, 2017.
\newblock ISSN 1662-5110.
\newblock \doi{10.3389/FNCIR.2017.00081}.

\bibitem[Heald et~al.(2021)Heald, Lengyel, and Wolpert]{Heald2021}
James~B Heald, M{\'{a}}t{\'{e}} Lengyel, and Daniel~M Wolpert.
\newblock {Contextual inference underlies the learning of sensorimotor
  repertoires}.
\newblock \emph{Nature}, 600\penalty0 (7889):\penalty0 489--493, 2021.
\newblock ISSN 1476-4687.
\newblock \doi{10.1038/s41586-021-04129-3}.

\bibitem[Holmgren et~al.(2003)Holmgren, Harkany, Svennenfors, and
  Zilberter]{holmgren2003}
C.~Holmgren, T.~Harkany, B.~Svennenfors, and Y.~Zilberter.
\newblock Pyramidal cell communication within local networks in layer 2/3 of
  rat neocortex.
\newblock \emph{The Journal of Physiology}, 551\penalty0 (1):\penalty0
  139--153, 2003.
\newblock ISSN 0022-3751.
\newblock \doi{10.1113/jphysiol.2003.044784}.

\bibitem[Hotelling(1931)]{hotelling1931}
Harold Hotelling.
\newblock The generalization of {S}tudent's ratio.
\newblock \emph{Annals of Mathematical Statistics}, 2\penalty0 (3):\penalty0
  360--378, 1931.
\newblock \doi{10.1007/978-1-4612-0919-5_4}.

\bibitem[Ibarz et~al.(2021)Ibarz, Tan, Finn, Kalakrishnan, Pastor, and
  Levine]{Ibarz2021}
Julian Ibarz, Jie Tan, Chelsea Finn, Mrinal Kalakrishnan, Peter Pastor, and
  Sergey Levine.
\newblock How to train your robot with deep reinforcement learning: lessons we
  have learned.
\newblock \emph{The International Journal of Robotics Research}, 40\penalty0
  (4-5):\penalty0 698--721, 2021.
\newblock \doi{10.1177/0278364920987859}.
\newblock URL \url{https://doi.org/10.1177/0278364920987859}.

\bibitem[Irpan(2018)]{Irpan2018}
Alex Irpan.
\newblock Deep reinforcement learning doesn't work yet, 2018.
\newblock URL \url{https://www.alexirpan.com/2018/02/14/rl-hard.html}.

\bibitem[Jadi et~al.(2014)Jadi, Behabadi, Poleg-Polsky, Schiller, and
  Mel]{jadi2014}
Monika~P. Jadi, Bardia~F. Behabadi, Alon Poleg-Polsky, Jackie Schiller, and
  Bartlett~W. Mel.
\newblock An augmented two-layer model captures nonlinear analog spatial
  integration effects in pyramidal neuron dendrites.
\newblock \emph{Proceedings of the IEEE, (Special issue on Computational
  Neuroscience)}, 102\penalty0 (5):\penalty0 782--798, 2014.
\newblock ISSN 0018-9219.
\newblock \doi{10.1109/JPROC.2014.2312671}.
\newblock URL \url{https://www.ncbi.nlm.nih.gov/pmc/articles/PMC4279447/}.

\bibitem[Jayakumar et~al.(2020)Jayakumar, Czarnecki, Menick, Schwarz, Rae,
  Osindero, Teh, Harley, and Pascanu]{jayakumar2019}
Siddhant~M. Jayakumar, Wojciech~M. Czarnecki, Jacob Menick, Jonathan Schwarz,
  Jack Rae, Simon Osindero, Yee~Whye Teh, Tim Harley, and Razvan Pascanu.
\newblock Multiplicative interactions and where to find them.
\newblock In \emph{Proceedings of the 8th International Conference on Learning
  Representations}, Digital, 2020.
\newblock URL \url{https://openreview.net/forum?id=rylnK6VtDH}.

\bibitem[Kandel(2012)]{kandel_principles_2012}
Eric Kandel.
\newblock \emph{Principles of Neural Science}.
\newblock McGraw-Hill, New York, NY, USA, 5 edition, 2012.
\newblock ISBN 0-07-139011-1.

\bibitem[Kendall et~al.(2019)Kendall, Gal, and Cipolla]{kendall2019multi}
Alex Kendall, Yarin Gal, and Roberto Cipolla.
\newblock Multi-task learning using uncertainty to weigh losses for scene
  geometry and semantics.
\newblock In \emph{Proceedings of the IEEE Conference on Computer Vision and
  Pattern Recognition}, Long Beach, USA, 2019.

\bibitem[Kerlin et~al.(2019)Kerlin, Boaz, Flickinger, Maclennan, Dean, Davis,
  Spruston, and Svoboda]{Kerlin2019}
Aaron Kerlin, Mohar Boaz, Daniel Flickinger, Bryan~J. Maclennan, Matthew~B.
  Dean, Courtney Davis, Nelson Spruston, and Karel Svoboda.
\newblock {Functional clustering of dendritic activity during decision-making}.
\newblock \emph{eLife}, 8, oct 2019.
\newblock \doi{10.7554/ELIFE.46966}.

\bibitem[Kingma and Ba(2015)]{kingma2015}
Diederik~P. Kingma and Jimmy~Lei Ba.
\newblock Adam: a method for stochastic optimization.
\newblock In \emph{Proceedings of the 3rd International Conference on Learning
  Representations}, San Diego, USA, 2015.

\bibitem[Kirkpatrick et~al.(2017)Kirkpatrick, Pascanu, Rabinowitz, Veness,
  Desjardins, Rusu, Milan, Quan, Ramalho, Grabska-Barwinska, Hassabis, Clopath,
  Kumaran, and Hadsell]{kirkpatrick2017}
James Kirkpatrick, Razvan Pascanu, Neil Rabinowitz, Joel Veness, Guillaume
  Desjardins, Andrei~A. Rusu, Kieran Milan, John Quan, Tiago Ramalho, Agnieszka
  Grabska-Barwinska, Demis Hassabis, Claudia Clopath, Dharshan Kumaran, and
  Raia Hadsell.
\newblock Overcoming catastrophic forgetting in neural networks.
\newblock \emph{Proceedings of the National Academy of Sciences}, 114\penalty0
  (13), 2017.
\newblock \doi{10.1073/pnas.1611835114}.

\bibitem[Lafourcade et~al.(2022)Lafourcade, {van der Goes}, Vardalaki, Brown,
  Voigts, Yun, Kim, Ku, and Harnett]{LAFOURCADE2022}
Mathieu Lafourcade, Marie-Sophie~H. {van der Goes}, Dimitra Vardalaki, Norma~J.
  Brown, Jakob Voigts, Dae~Hee Yun, Minyoung~E. Kim, Taeyun Ku, and Mark~T.
  Harnett.
\newblock Differential dendritic integration of long-range inputs in
  association cortex via subcellular changes in synaptic ampa-to-nmda receptor
  ratio.
\newblock \emph{Neuron}, 2022.
\newblock ISSN 0896-6273.
\newblock \doi{https://doi.org/10.1016/j.neuron.2022.01.025}.
\newblock URL
  \url{https://www.sciencedirect.com/science/article/pii/S0896627322000642}.

\bibitem[Lapique(1907)]{lapicque1907}
Louis Lapique.
\newblock Recherches quantitatives sur l'excitation \'{e}lectrique des nerfs
  trait\'{e}e comme une polarisation.
\newblock \emph{Journal of Physiology and Pathololgy}, 9:\penalty0 620--635,
  1907.

\bibitem[Larkum(2022)]{LARKUM2022}
Matthew Larkum.
\newblock {Are dendrites conceptually useful?}
\newblock \emph{Neuroscience}, 2022.
\newblock ISSN 0306-4522.
\newblock \doi{https://doi.org/10.1016/j.neuroscience.2022.03.008}.
\newblock URL
  \url{https://www.sciencedirect.com/science/article/pii/S0306452222001208}.

\bibitem[Larkum et~al.(1999)Larkum, Zhu, and Sakmann]{larkum_new_1999}
Matthew~E. Larkum, J.~Julius Zhu, and Bert Sakmann.
\newblock A new cellular mechanism for coupling inputs arriving at different
  cortical layers.
\newblock \emph{Nature}, 398\penalty0 (6725):\penalty0 338--341, 1999.
\newblock ISSN 1476-4687.
\newblock \doi{10.1038/18686}.
\newblock URL \url{https://www.nature.com/articles/18686}.

\bibitem[LeCun et~al.(2015)LeCun, Bengio, and Hinton]{lecun_deep_2015}
Yann LeCun, Yoshua Bengio, and Geoffrey Hinton.
\newblock Deep learning.
\newblock \emph{Nature}, 521\penalty0 (7553):\penalty0 436--444, 2015.
\newblock ISSN 1476-4687.
\newblock \doi{10.1038/nature14539}.
\newblock URL \url{https://www.nature.com/articles/nature14539}.

\bibitem[Liang et~al.(2019)Liang, Li, Chou, Zhou, Zhang, Xiao, Zhang, Tao, and
  Zhang]{liang_sparse_2019}
Feixue Liang, Haifu Li, Xiao-lin Chou, Mu~Zhou, Nicole~K Zhang, Zhongju Xiao,
  Ke~K Zhang, Huizhong~W Tao, and Li~I Zhang.
\newblock Sparse representation in awake auditory cortex: Cell-type dependence,
  synaptic mechanisms, developmental emergence, and modulation.
\newblock \emph{Cerebral Cortex}, 29\penalty0 (9):\penalty0 3796--3812, 2019.
\newblock ISSN 1047-3211.
\newblock \doi{10.1093/cercor/bhy260}.
\newblock URL \url{https://doi.org/10.1093/cercor/bhy260}.

\bibitem[Lillicrap et~al.(2016)Lillicrap, Hunt, Pritzel, Heess, Erez, Tassa,
  Silver, and Wierstra]{lillicrap2016continuous}
Timothy~P. Lillicrap, Jonathan~J. Hunt, Alexander Pritzel, Nicolas Heess, Tom
  Erez, Yuval Tassa, David Silver, and Daan Wierstra.
\newblock Continuous control with deep reinforcement learning.
\newblock In \emph{Proceedings of the 4th International Conference on Learning
  Representations}, San Juan, Puerto Rico, 2016.

\bibitem[Limbacher and Legenstein(2020)]{limbacher_emergence_2020}
Thomas Limbacher and Robert Legenstein.
\newblock Emergence of stable synaptic clusters on dendrites through synaptic
  rewiring.
\newblock \emph{Frontiers in Computational Neuroscience}, 14\penalty0
  (57):\penalty0 1--19, 2020.
\newblock ISSN 1662-5188.
\newblock \doi{10.3389/fncom.2020.00057}.
\newblock URL
  \url{https://www.frontiersin.org/article/10.3389/fncom.2020.00057}.

\bibitem[Liu et~al.(2019)Liu, Johns, and Davison]{liu2019endtoend}
Shikun Liu, Edward Johns, and Andrew~J. Davison.
\newblock End-to-end multi-task learning with attention.
\newblock In \emph{Proceedings of the IEEE Conference on Computer Vision and
  Pattern Recognition}, Long Beach, USA, 2019.

\bibitem[London and H{\"{a}}usser(2005)]{london_dendritic_2005}
Michael London and Michael H{\"{a}}usser.
\newblock Dendritic computation.
\newblock \emph{Annual Review of Neuroscience}, 28\penalty0 (1):\penalty0
  503--532, 2005.
\newblock \doi{10.1146/annurev.neuro.28.061604.135703}.

\bibitem[Losonczy et~al.(2008)Losonczy, Makara, and Magee]{losonczy2008}
Attila Losonczy, Judit~K. Makara, and Jeffrey~C. Magee.
\newblock Compartmentalized dendritic plasticity and input feature storage in
  neurons.
\newblock \emph{Nature}, 452\penalty0 (7186):\penalty0 436--441, 2008.
\newblock \doi{10.1038/nature06725}.

\bibitem[Magee(2000)]{Magee2000}
Jeffrey~C Magee.
\newblock {Dendritic integration of excitatory synaptic input}.
\newblock \emph{Nature Reviews Neuroscience}, 1\penalty0 (3):\penalty0
  181--190, 2000.
\newblock ISSN 1471-0048.
\newblock \doi{10.1038/35044552}.
\newblock URL \url{https://doi.org/10.1038/35044552}.

\bibitem[Majani et~al.(1989)Majani, Erlanson, and Abu-Mostafa]{majani1989}
E.~Majani, R.~Erlanson, and Y.~Abu-Mostafa.
\newblock On the k-winners-take-all network.
\newblock In \emph{Advances in Neural Information Processing Systems}, Denver,
  USA, 1989.

\bibitem[Major et~al.(2013)Major, Larkum, and Schiller]{major2013}
Guy Major, Matthew~E. Larkum, and Jackie Schiller.
\newblock Active properties of neocortical pyramidal neuron dendrites.
\newblock \emph{Annual Review of Neuroscience}, 36:\penalty0 1--24, 2013.
\newblock \doi{10.1146/annurev-neuro-062111-150343}.

\bibitem[Maninis et~al.(2019)Maninis, Radosavovic, and
  Kokkinos]{maninis2019attentive}
Kevis-Kokitsi Maninis, Ilija Radosavovic, and Iasonas Kokkinos.
\newblock Attentive single-tasking of multiple tasks.
\newblock In \emph{Proceedings of the IEEE Conference on Computer Vision and
  Pattern Recognition}, Long Beach, USA, 2019.

\bibitem[Masse et~al.(2018)Masse, Grant, and Freedman]{masse2018}
Nicolas~Y. Masse, Gregory~D. Grant, and David~J. Freedman.
\newblock Alleviating catastrophic forgetting using context-dependent gating
  and synaptic stabilization.
\newblock \emph{Proceedings of the National Academy of Sciences}, 115\penalty0
  (44), 2018.
\newblock \doi{10.1073/pnas.1803839115}.

\bibitem[McCann et~al.(2018)McCann, Keskar, Xiong, and
  Socher]{mccann2018natural}
Bryan McCann, Nitish~Shirish Keskar, Caiming Xiong, and Richard Socher.
\newblock The natural language decathlon: Multitask learning as question
  answering.
\newblock \emph{arXiv:1806.08730}, 2018.

\bibitem[McClelland et~al.(1986)McClelland, Rumelhart, and the PDP
  Research~Group]{mcclelland1986parallel}
James~L. McClelland, David~E. Rumelhart, and the PDP Research~Group.
\newblock \emph{Parallel Distributed Processing}.
\newblock MIT Press, Cambridge, MA, USA, 2 edition, 1986.

\bibitem[McCloskey and Cohen(1989)]{mccloskey1989}
Michael McCloskey and Neal~J. Cohen.
\newblock Catastrophic interference in connectionist networks: The sequential
  learning problem.
\newblock \emph{Psychology of Learning and Motivation - Advances in Research
  and Theory}, 24\penalty0 (C):\penalty0 109--165, 1989.
\newblock ISSN 0079-7421.
\newblock \doi{10.1016/S0079-7421(08)60536-8}.

\bibitem[Misra et~al.(2016)Misra, Shrivastava, Gupta, and
  Hebert]{misra2016crossstitch}
Ishan Misra, Abhinav Shrivastava, Abhinav Gupta, and Martial Hebert.
\newblock Cross-stitch networks for multi-task learning.
\newblock In \emph{Proceedings of the IEEE Conference on Computer Vision and
  Pattern Recognition}, Las Vegas, USA, 2016.

\bibitem[Mnih et~al.(2013)Mnih, Kavukcuoglu, Silver, Graves, Antonoglou,
  Wierstra, and Riedmiller]{mnih2013playing}
Volodymyr Mnih, Koray Kavukcuoglu, David Silver, Alex Graves, Ioannis
  Antonoglou, Daan Wierstra, and Martin Riedmiller.
\newblock Playing atari with deep reinforcement learning.
\newblock In \emph{Advances in Neural Information Processing Systems}, Lake
  Tahoe, USA, 2013.

\bibitem[Paiton et~al.(2020)Paiton, Frye, Lundquist, Bowen, Zarcone, and
  Olshausen]{Paiton2020}
Dylan~M Paiton, Charles~G Frye, Sheng~Y Lundquist, Joel~D Bowen, Ryan Zarcone,
  and Bruno~A Olshausen.
\newblock {Selectivity and robustness of sparse coding networks}.
\newblock \emph{Journal of Vision}, 20\penalty0 (12):\penalty0 10, 2020.
\newblock ISSN 1534-7362.
\newblock \doi{10.1167/jov.20.12.10}.

\bibitem[Parisi et~al.(2019)Parisi, Kemker, Part, Kanan, and
  Wermter]{parisi_continual_2019}
German~I. Parisi, Ronald Kemker, Jose~L. Part, Christopher Kanan, and Stefan
  Wermter.
\newblock Continual lifelong learning with neural networks: {A} review.
\newblock \emph{Neural Networks}, 113:\penalty0 54--71, 2019.
\newblock ISSN 0893-6080.
\newblock \doi{10.1016/j.neunet.2019.01.012}.
\newblock URL
  \url{https://www.sciencedirect.com/science/article/pii/S0893608019300231}.

\bibitem[Phillips et~al.(2015)Phillips, Clark, and Silverstein]{Phillips2015a}
W.~A. Phillips, A.~Clark, and S.~M. Silverstein.
\newblock {On the functions, mechanisms, and malfunctions of intracortical
  contextual modulation}.
\newblock \emph{Neuroscience and Biobehavioral Reviews}, 52\penalty0
  (February):\penalty0 1--20, 2015.
\newblock ISSN 18737528.
\newblock \doi{10.1016/j.neubiorev.2015.02.010}.

\bibitem[Phillips(2015)]{Phillips2015}
William~A. Phillips.
\newblock {Cognitive functions of intracellular mechanisms for contextual
  amplification}.
\newblock \emph{Brain and Cognition}, 2015.
\newblock ISSN 10902147.
\newblock \doi{10.1016/j.bandc.2015.09.005}.

\bibitem[Poirazi and Papoutsi(2020)]{poirazi2020}
Panayiota Poirazi and Athanasia Papoutsi.
\newblock Illuminating dendritic function with computational models.
\newblock \emph{Nature Reviews Neuroscience}, 21:\penalty0 303--321, 2020.
\newblock \doi{10.1038/s41583-020-0301-7}.

\bibitem[Poirazi et~al.(2003)Poirazi, Brannon, and Mel]{poirazi2003}
Panayiota Poirazi, Terrence Brannon, and Bartlett~W. Mel.
\newblock Pyramidal neuron as two-layer neural network.
\newblock \emph{Neuron}, 37\penalty0 (6):\penalty0 989--999, 2003.
\newblock ISSN 0896-6273.
\newblock \doi{10.1016/S0896-6273(03)00149-1}.
\newblock URL
  \url{https://www.sciencedirect.com/science/article/pii/S0896627303001491}.

\bibitem[Purushwalkam et~al.(2019)Purushwalkam, Nickel, Gupta, and
  Ranzato]{purushwalkam2019task}
Senthil Purushwalkam, Maximilian Nickel, Abhinav Gupta, and Marc'Aurelio
  Ranzato.
\newblock Task-driven modular networks for zero-shot compositional learning.
\newblock In \emph{Proceedings of the IEEE International Conference on Computer
  Vision}, Long Beach, USA, 2019.

\bibitem[Ramaswamy and Markram(2015)]{Ramaswamy2015}
Srikanth Ramaswamy and Henry Markram.
\newblock {Anatomy and physiology of the thick-tufted layer 5 pyramidal
  neuron.}
\newblock \emph{Frontiers in cellular neuroscience}, 9\penalty0 (June), 2015.
\newblock ISSN 1662-5102.
\newblock \doi{10.3389/fncel.2015.00233}.

\bibitem[Rosch(1975)]{rosch1975}
Eleanor Rosch.
\newblock Cognitive representations of semantic categories.
\newblock \emph{Journal of Experimental Psychology: {G}eneral}, 104:\penalty0
  192--233, 1975.
\newblock \doi{0096-3445.104.3.192}.

\bibitem[Rosenbaum et~al.(2018)Rosenbaum, Klinger, and
  Riemer]{rosenbaum2018routing}
Clemens Rosenbaum, Tim Klinger, and Matthew Riemer.
\newblock Routing networks: Adaptive selection of non-linear functions for
  multi-task learning.
\newblock In \emph{Proceedings of the 6th International Conference on Learning
  Representations}, Vancouver, Canada, 2018.
\newblock URL \url{https://openreview.net/forum?id=ry8dvM-R-}.

\bibitem[Rosenblatt(1958)]{rosenblatt1958}
Frank Rosenblatt.
\newblock The perceptron: a probabilistic model for information storage and
  organization in the brain.
\newblock \emph{Psychological review}, 65\penalty0 (6):\penalty0 386, 1958.
\newblock \doi{10.1037/h0042519}.

\bibitem[Rumelhart et~al.(1986)Rumelhart, Hinton, and Williams]{rumelhart1986}
David~E. Rumelhart, Geoffrey~E. Hinton, and Ronald~J. Williams.
\newblock Learning representations by back-propagating errors.
\newblock \emph{Nature}, 323:\penalty0 533--536, 1986.
\newblock \doi{10.1038/323533a0}.

\bibitem[Rusu et~al.(2016)Rusu, Colmenarejo, G\"{u}l\c{c}ehre, Desjardins,
  Kirkpatrick, Pascanu, Mnih, Kavukcuoglu, and Hadsell]{rusu2016policy}
Andrei~A. Rusu, Sergio~G\'{o}mez Colmenarejo, Caglar G\"{u}l\c{c}ehre,
  Guillaume Desjardins, James Kirkpatrick, Razvan Pascanu, Volodymyr Mnih,
  Koray Kavukcuoglu, and Raia Hadsell.
\newblock Policy distillation.
\newblock In \emph{Proceedings of the 4th International Conference on Learning
  Representations}, San Juan, Puerto Rico, 2016.

\bibitem[Sahni et~al.(2017)Sahni, Kumar, Tejani, and Isbell]{sahni2017learning}
Himanshu Sahni, Saurabh Kumar, Farhan Tejani, and Charles Isbell.
\newblock Learning to compose skills.
\newblock In \emph{Advances in Neural Information Processing Systems}, Long
  Beach, USA, 2017.

\bibitem[Schmidt-Hieber et~al.(2017)Schmidt-Hieber, Toleikyte, Aitchison, Roth,
  Clark, Branco, and H{\"{a}}usser]{Schmidt-Hieber2017}
Christoph Schmidt-Hieber, Gabija Toleikyte, Laurence Aitchison, Arnd Roth,
  Beverley~A Clark, Tiago Branco, and Michael H{\"{a}}usser.
\newblock {Active dendritic integration as a mechanism for robust and precise
  grid cell firing}.
\newblock \emph{Nature Neuroscience}, \penalty0 (June), 2017.
\newblock ISSN 1097-6256.
\newblock \doi{10.1038/nn.4582}.

\bibitem[Schulman et~al.(2017)Schulman, Wolski, Dhariwal, Radford, and
  Klimov]{schulman2017proximal}
John Schulman, Filip Wolski, Prafulla Dhariwal, Alec Radford, and Oleg Klimov.
\newblock Proximal policy optimization algorithms.
\newblock \emph{arXiv:1707.06347}, 2017.

\bibitem[Sener and Koltun(2018)]{sener2019multitask}
Ozan Sener and Vladlen Koltun.
\newblock Multi-task learning as multi-objective optimization.
\newblock In \emph{Advances in Neural Information Processing Systems},
  Montr\'{e}al, Canada, 2018.

\bibitem[Sezener et~al.(2021)Sezener, Grabska-Barwi\'{n}ska, Kostadinov, Beau,
  Krishnagopal, Budden, Hutter, Veness, Botvinick, Clopath, H\"{a}usser, and
  Latham]{sezener2021}
Eren Sezener, Agnieszka Grabska-Barwi\'{n}ska, Dimitar Kostadinov, Maxime Beau,
  Sanjukta Krishnagopal, David Budden, Marcus Hutter, Joel Veness, Matthew
  Botvinick, Claudia Clopath, Michael H\"{a}usser, and Peter~E. Latham.
\newblock A rapid and efficient learning rule for biological neural circuits.
\newblock \emph{bioRxiv}, 2021.
\newblock \doi{10.1101/2021.03.10.434756}.

\bibitem[Siegel et~al.(2000)Siegel, K{\"{o}}rding, and K{\"{o}}nig]{Siegel2000}
Markus Siegel, Konrad~P. K{\"{o}}rding, and Peter K{\"{o}}nig.
\newblock {Integrating top-down and bottom-up sensory processing by
  somato-dendritic interactions}.
\newblock \emph{Journal of Computational Neuroscience}, 8\penalty0
  (2):\penalty0 161--173, 2000.
\newblock ISSN 09295313.
\newblock \doi{10.1023/A:1008973215925}.

\bibitem[Silver et~al.(2018)Silver, Hubert, Schrittwieser, Antonoglou, Lai,
  Guez, Lanctot, Sifre, Kumaran, Graepel, Lillicrap, Simonyan, and
  Hassabis]{silver2018general}
David Silver, Thomas Hubert, Julian Schrittwieser, Ioannis Antonoglou, Matthew
  Lai, Arthur Guez, Marc Lanctot, Laurent Sifre, Dharshan Kumaran, Thore
  Graepel, Timothy Lillicrap, Karen Simonyan, and Demis Hassabis.
\newblock A general reinforcement learning algorithm that masters chess, shogi,
  and go through self-play.
\newblock \emph{Science}, 362\penalty0 (6419):\penalty0 1140--1144, 2018.
\newblock \doi{10.1126/science.aar6404}.

\bibitem[Snell et~al.(2017)Snell, Swersky, and Zemel]{snell2017}
Jake Snell, Kevin Swersky, and Richard~S. Zemel.
\newblock Prototypical networks for few-shot learning.
\newblock In \emph{Advances in Neural Information Processing Systems}, Long
  Beanch, USA, 2017.

\bibitem[Spruston(2008)]{spruston2008}
Nelson Spruston.
\newblock Pyramidal neurons: dendritic structure and synaptic integration.
\newblock \emph{Nature Reviews Neuroscience}, 9:\penalty0 206--221, 2008.
\newblock ISSN 1471-003X.
\newblock \doi{10.1038/nrn2286}.

\bibitem[Stuart et~al.(2016)Stuart, Spruston, and H{\"{a}}usser]{Stuart2016}
G~Stuart, Nelson Spruston, and M~H{\"{a}}usser, editors.
\newblock \emph{{Dendrites}}.
\newblock Oxford University Press, USA, Oxford, UK, third edition, 2016.
\newblock ISBN 978–0–19–874527–3.

\bibitem[Stuart and Spruston(2015)]{Stuart2015}
Greg~J Stuart and Nelson Spruston.
\newblock {Dendritic integration: 60 years of progress}.
\newblock \emph{Nature Neuroscience}, 2015.
\newblock ISSN 1097-6256.
\newblock \doi{10.1038/nn.4157}.

\bibitem[Sutton and Barto(2018)]{sutton2018reinforcement}
Richard~S. Sutton and Andrew~G. Barto.
\newblock \emph{Reinforcement Learning: An Introduction}.
\newblock MIT Press, Cambridge, MA, USA, 2 edition, 2018.
\newblock URL \url{http://incompleteideas.net/book/the-book-2nd.html}.

\bibitem[Takahashi et~al.(2020)Takahashi, Ebner, Sigl-Gl{\"{o}}ckner, Moberg,
  Nierwetberg, and Larkum]{takahashi2020}
Naoya Takahashi, Christian Ebner, Johanna Sigl-Gl{\"{o}}ckner, Sara Moberg,
  Svenja Nierwetberg, and Matthew~E. Larkum.
\newblock Active dendritic currents gate descending cortical outputs in
  perception.
\newblock \emph{Nature Neuroscience}, 23:\penalty0 1--9, 2020.
\newblock ISSN 1097-6256.
\newblock \doi{10.1038/s41593-020-0677-8}.
\newblock URL \url{http://www.nature.com/articles/s41593-020-0677-8}.

\bibitem[van~de Ven and Tolias(2019)]{van_de_ven_three_2019}
Gido~M. van~de Ven and Andreas~S. Tolias.
\newblock Three scenarios for continual learning.
\newblock \emph{arXiv:1904.07734}, 2019.
\newblock URL \url{http://arxiv.org/abs/1904.07734}.

\bibitem[Veness et~al.(2021)Veness, Lattimore, Budden, Bhoopchand, Mattern,
  Grabska-Barwinska, Sezener, Wang, Toth, Schmitt, and Hutter]{veness2021}
Joel Veness, Tor Lattimore, David Budden, Avishkar Bhoopchand, Christopher
  Mattern, Agnieszka Grabska-Barwinska, Eren Sezener, Jianan Wang, Peter Toth,
  Simon Schmitt, and Marcus Hutter.
\newblock Gated linear networks.
\newblock In \emph{Proceedings of the 35th AAAI Conference on Artificial
  Intelligence}, Digital, 2021.

\bibitem[Whittington et~al.(2022)Whittington, McCaffary, Bakermans, and
  Behrens]{Whittington2022}
James C.~R. Whittington, David McCaffary, Jacob J.~W. Bakermans, and Timothy
  E.~J. Behrens.
\newblock {How to build a cognitive map: insights from models of the
  hippocampal formation}.
\newblock feb 2022.
\newblock \doi{10.48550/arxiv.2202.01682}.
\newblock URL \url{https://arxiv.org/abs/2202.01682}.

\bibitem[Wilson et~al.(2007)Wilson, Fern, Ray, and Tadepalli]{wilson2007multi}
Aaron Wilson, Alan Fern, Soumya Ray, and Prasad Tadepalli.
\newblock Multi-task reinforcement learning: A hierarchical bayesian approach.
\newblock In \emph{Proceedings of the 24th International Conference on Machine
  Learning}, Corvallis, USA, 2007.

\bibitem[Wortsman et~al.(2020)Wortsman, Ramanujan, Liu, Kembhavi, Rastegari,
  Yosinski, and Farhadi]{wortsman2020}
Mitchell Wortsman, Vivek Ramanujan, Rosanne Liu, Aniruddha Kembhavi, Mohammad
  Rastegari, Jason Yosinski, and Ali Farhadi.
\newblock Supermasks in superposition.
\newblock In \emph{Advances in Neural Information Processing Systems}, Digital,
  2020.

\bibitem[y~Cajal(1894)]{cajal1894}
Santiago~Ram{\'o}n y~Cajal.
\newblock Neue darstellung vom histologischen bau des centralnervensystems.
\newblock \emph{American Journal of Psychology}, 6\penalty0 (3):\penalty0 450,
  1894.
\newblock \doi{10.2307/1411662}.

\bibitem[Yang et~al.(2014)Yang, Lai, Cichon, Ma, Li, and Gan]{yang2014}
Guang Yang, Cora Sau~Wan Lai, Joseph Cichon, Lei Ma, Wei Li, and Wen-Biao Gan.
\newblock Sleep promotes branch-specific formation of dendritic spines after
  learning.
\newblock \emph{Science}, 344\penalty0 (6188):\penalty0 1173--1178, 2014.
\newblock ISSN 1095-9203.
\newblock \doi{10.1126/science.1249098}.
\newblock URL \url{http://www.ncbi.nlm.nih.gov/pubmed/24904169}.

\bibitem[Yang et~al.(2020)Yang, Xu, WU, and Wang]{yang2020multi}
Ruihan Yang, Huazhe Xu, YI~WU, and Xiaolong Wang.
\newblock Multi-task reinforcement learning with soft modularization.
\newblock In \emph{Advances in Neural Information Processing Systems}, Digital,
  2020.
\newblock URL
  \url{https://proceedings.neurips.cc/paper/2020/file/32cfdce9631d8c7906e8e9d6e68b514b-Paper.pdf}.

\bibitem[Yu et~al.(2019)Yu, Quillen, He, Julian, Narayan, Shively, Bellathur,
  Hausman, Finn, and Levine]{yu2019}
Tianhe Yu, Deirdre Quillen, Zhanpeng He, Ryan Julian, Avnish Narayan, Hayden
  Shively, Adithya Bellathur, Karol Hausman, Chelsea Finn, and Sergey Levine.
\newblock Meta-world: A benchmark and evaluation for multi-task and meta
  reinforcement learning.
\newblock In \emph{Proceedings of the 3rd Conference on Robot Learning}, Osaka,
  Japan, 2019.

\bibitem[Yu et~al.(2020)Yu, Kumar, Gupta, Levine, Hausman, and
  Finn]{yu2020gradient}
Tianhe Yu, Saurabh Kumar, Abhishek Gupta, Sergey Levine, Karol Hausman, and
  Chelsea Finn.
\newblock Gradient surgery for multi-task learning.
\newblock In \emph{Advances in Neural Information Processing Systems}, 2020.

\bibitem[Zenke et~al.(2017)Zenke, Poole, and Ganguli]{zenke2017}
Friedemann Zenke, Ben Poole, and Surya Ganguli.
\newblock Continual learning through synaptic intelligence.
\newblock In \emph{Proceedings of the 34th International Conference on Machine
  Learning}, Digital, 2017.

\bibitem[Zhang and Yeung(2014)]{zhang2014regularization}
Yu~Zhang and Dit-Yan Yeung.
\newblock A regularization approach to learning task relationships in multitask
  learning.
\newblock \emph{ACM Transactions on Knowledge Discovery from Data}, 8\penalty0
  (3):\penalty0 1--31, 2014.
\newblock ISSN 1556-4681.
\newblock \doi{10.1145/2538028}.
\newblock URL \url{https://doi.org/10.1145/2538028}.

\end{thebibliography}

%%% Make sure to upload the bib file along with the tex file and PDF
%%% Please see the test.bib file for some examples of references
\appendix
\addtocontents{toc}{\protect\contentsline{chapter}{Appendix:}{}}

\newpage

\section*{Appendix}

\section{Multitask Reinforcement Learning}
\label{section:additional_mtrl_experiments}
\subsection{Additional Experiments}

We ran a number of control experiments that test variations of our Active Dendrites Network in order to better understand the impact of architectural decisions (Table~\ref{table:additional_mtrl_hyperparameters}). For these experiments we use the hyperparameters as presented in Table~\ref{table:mtrl_hyperparameters}. The average success rate for each architecture is averaged over 5 independent trials across the last 500,000 environment steps of training. For comparison, the Active Dendrites Network we present in Table~\ref{table:mtrl_hyperparameters} achieved an average success rate of \bacc{87.5}.

\begin{table*}[h]
\renewcommand*{\arraystretch}{1.3}
\vskip 0.15in

\begin{center}
\begin{tabular}{ l c c c c c}
\toprule
 & 1-Layer & 3-Layer & 1-Seg. & 20-Seg. & 2-Modulated\\
\midrule
\multicolumn{5}{l}{Network Details}\\
\bottomrule
Hidden Sizes & 1$\times$[4,000] & 3$\times$[2,000] & 2$\times$[2,800] & 2$\times$[2,800] & 2$\times$[2,800]\\
\\
Num. Dendritic Segments & 10 & 10 & 1 & 20 & 10\\
\\
Hidden Layers Modulated & [$\checkmark$] & [$\times$, $\times$, $\checkmark$] & [$\times$, $\checkmark$] & [$\times$, $\checkmark$] & [$\checkmark$, $\checkmark$]\\
\\
Total Non-zero Parameters & 558,804 & 7,483,404 & 7,197,964 & 7,729,964 & 7,729,964\\
\\
Average Success Rate & \bacc{58.9} & \bacc{87.7} & \bacc{85.6} & \bacc{89} & \bacc{82.5} \\
\bottomrule
\end{tabular}
\end{center}
\vskip -0.1in
\caption{The architecture details for each additional multi-task RL experiment. The \textbf{\textit{n}-Layer} experiments correspond to networks with \textbf{\textit{n}} hidden layers and modulated by 10 dendrites in the final hidden layer. The \textbf{\textit{n}-Seg.} experiments correspond to networks with 2 hidden layers and modulated by \textbf{\textit{n}} dendrites in the final hidden layer. The \textbf{2-Modulated} experiment uses a 2 hidden layer network, where both hidden layers are modulated by 10 dendrites each.}
\label{table:additional_mtrl_hyperparameters}
\end{table*}

Our Active Dendrites Network experiments suggest that using more than a single hidden layer and modulating only the final hidden layer produces the best results. In addition, modulation with more dendritic segments than the number of tasks results in marginal increases in success, while less dendritic segments than the number of tasks produces a marginal decrease in success. These findings are consistent with those observed in our continual learning experiments. 

\section{Continual Learning}
\subsection{Understanding Parameters in the Model}
\label{section:understanding_continual_learning_parameters}

One interesting issue relates to the size of the Active Dendrites Network and the total number of non-zero parameters. In addition to feedforward weights, our neurons have weights associated with each dendritic segment. In most of our experiments the number of dendritic segments is set to $\mathcal{T}$, the number of tasks (see Appendix~\ref{section:ten_dendrites} for results with a fixed number of dendritic segments). We can calculate the number of weights in each hidden layer $l$ of the network as follows. Let $p$ be the size of the prototype vector, $n_l$ be the number of units in layer $l$, $s^F$ be the weight sparsity for the feedforward weights and $s^D$ the weight sparsity for dendritic weights. The total number of weights in layer $l$ is then:

$$W_l = \underbrace{\left( \left( 1 - s^F \right)  n_l + 1 \right)  n_{l-1}}_{\text{Feedforward weights \& biases}} + \underbrace{\left( 1 - s^D \right) p \mathcal{T} n_l}_{\text{Dendritic segments}}$$

The first term represents the total number of weights in the feedforward portion (including a bias). The second term represents the number of weights in the dendritic segments. In our implementation, $s^F = 0.5$, and $s^D = 0$ (i.e., dendritic weights are fully dense).

In addition to these weights we also store $\mathcal{T}$ prototypes, each of which has the same size as the input vector. Although these are not learned through backpropagation, they are determined from the training data and should be included in the parameter count. In permutedMNIST the input vector size is $n_0=784$, leading to a total of $\mathcal{T} \times 784$ additional values for the prototypes. 

The number of dendritic weights quickly dominates all other parameters as the number of tasks increases (Table~\ref{table:additional_cl_hyperparameters}, middle column). At first glance, the implication is that the number of parameters in our 100-task network is far greater than the number of parameters in the comparison networks. However notice that the dendritic segments do not receive the input. The dendritic segments determine a context-dependent scale factor per neuron, based only on one of $\mathcal{T}$ possible context vectors. This scale factor is learned during training but then is static during testing.

Since there is a small fixed pool of $\mathcal{T}$ prototype vectors, a simple post-processing step can replace the weights with a smaller identical system.
The output of an Active Dendrites Neuron is:

$$\hat{y} = \left( \w^\top \x + b \right) \times \sigma \left( \max_j \u_j^\top \context \right)$$

\noindent During testing $\sigma \left( \max_j \u_j^\top \context \right)$ is constant for each vector $\context_i$. The equation can be re-written as:

$$
\hat{y} = \left( \w^\top \x + b \right) d_i
$$

\noindent where $d_i = \sigma \left( \max_j \u_j^\top \context_i \right)$, $0<i \leq \mathcal{T}$. For any given test input, we can select the nearest prototype vector $i$ and use the appropriate scale factor $d_i$. The total number of {\it effective parameters} in layer $l$ is thus reduced to:

$$W'_l = \left( \left( 1 - s^F \right) n_l + 1 \right) n_{l-1} + \mathcal{T} n_l$$

Note that in the experiments reported here, with a small fixed number of context vectors, it is actually possible to learn $d_i$ directly via backpropagation. In this case the number of non-zero parameters would be identical to the number of effective parameters, even during training. We did not implement this as it would also limit the flexibility of the overall architecture and disallow future scenarios where the context vector changes dynamically per input. In Table~\ref{table:additional_cl_hyperparameters} we list the total and effective number of parameters for the Active Dendrites Network in comparison to some of the other networks. Note that the Active Dendrites Network has substantially fewer effective parameters than any of the other networks.

\begin{table*}[ht]
\vskip 0.15in
\begin{center}
\begin{small}
% \begin{sc}
\begin{tabular}{lccc}
\toprule
 Network       & Tasks & Non-Zero Parameters & Effective Parameters \\
\midrule
 Active Dendrites Network     & 10 & 35,034,794 & 2,963,114 \\
 Active Dendrites Network     & 100 & 324,119,114 & 3,402,314 \\
 % NOTE: Karan increased the parameter count for 3- and 10-layer MLPs by 10 on 12/21/21 since they did not count for output layer biases.
 3-layer MLP                  & 10 or 100 & 5,824,522 & 5,824,522 \\
 10-layer MLP                 & 10 or 100 & 35,198,986 & 35,198,986 \\
 XdG                          & 10 or 100 & 5,592,010 & 5,592,010 \\ 
\bottomrule
\end{tabular}
% \end{sc}
\end{small}
\end{center}
\vskip -0.1in
\caption{The total and effective number of parameters for Active Dendrites Networks as compared to some of the other networks. Note: that XdG requires a mapping from task ID to sub-networks that is not incorporated in the table.}
\label{table:additional_cl_hyperparameters}
\end{table*}

In our previous work \citep{hawkins2016} we used extremely sparse dendritic weights ($>99\%$ sparsity). These weights were dynamically determined during the learning process by sampling from components of the context vector. Consistent with the biology of active dendrites, the number of weights per segment was limited to a small constant (such as $30$). Implementing sparse dendritic weights in the context of deep learning systems is an important future research area for Active Dendrites Networks. \newline

\subsection{Number of Clusters Formed when Inferring Prototypes}

\begin{figure}[ht]
\centering
\includegraphics[width=3.0in]{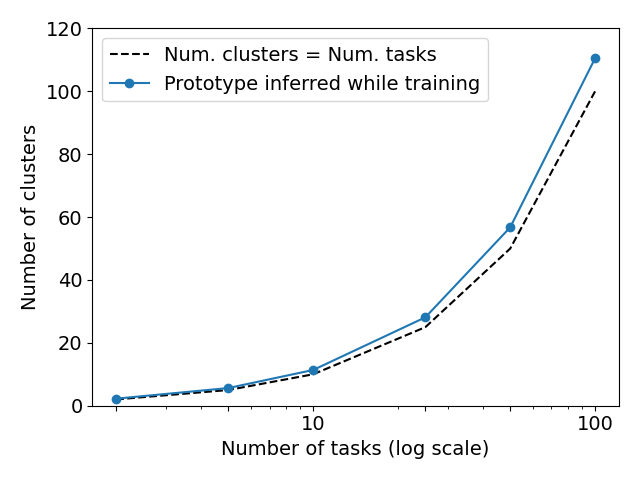}
\caption{
The average number of clusters found by the clustering procedure (described by Algorithms~\ref{alg:cluster} and~\ref{alg:hotelling} as a function of the number of permutedMNIST tasks.
All results are averaged over 8 independent trials.
}
\label{fig:num_clusters}
\end{figure}

In our continual learning experiments, when inferring the context vector while training via clustering, we used a significance threshold of $p = 0.9$.
We chose this value arbitrarily, and can further improve our results by incorporating $p$ as a model hyperparameter.
Assuming that the prototype vector for each permutedMNIST task is sufficiently different, we found that our method arrives at a ``sensible'' number of prototypes (i.e., not too few nor too many clusters/prototypes as compared with the number of tasks).
Figure \ref{fig:num_clusters} shows the average number of clusters formed as a function of the number of permutedMNIST tasks we trained an Active Dendrites Network on.

\subsection{Active Dendrites Network with a Fixed Number of Parameters}
\label{section:ten_dendrites}

\begin{figure}[ht]
\centering
\includegraphics[width=3.0in]{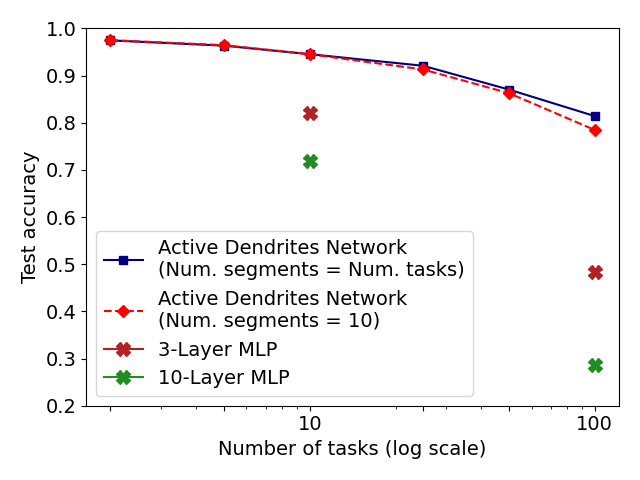}
\caption{
Continual learning accuracy on permutedMNIST: an Active Dendrites Network with the same number of dendritic segments per neuron as the number of tasks (blue), and one with exactly 10 dendritic segments per neuron (red).
We also included 3- and 10-layer MLPs on 10 and 100 tasks.
All results are averaged over 8 independent trials.
}
\label{fig:10_dendrites}
\end{figure}

In our continual learning experiments we mentioned that for any given number of permutedMNIST tasks, a single Active Dendrites Neuron has the same number of dendritic segments as tasks.
The total number of learnable, non-zero parameters in that scenario grows linearly with the number of tasks.
(Table~\ref{table:additional_cl_hyperparameters} lists each model's parameter count.)
Although the number of effective parameters is far smaller than the actual parameter count (see Appendix~\ref{section:understanding_continual_learning_parameters}), we also tested learning 100 tasks in sequence with a fixed 10 dendritic segments per neuron.
This network maintains a constant 35 million non-zero parameters (same as a 10-layer MLP) independent of the number of tasks.
As Figure \ref{fig:10_dendrites} shows, our modified network achieves $78.5\%$ accuracy on 100 tasks, close to the network with 100 dendritic segments.

Why does an Active Dendrites Network not suffer from a severe drop in accuracy with significantly fewer dendritic segments for a large number of tasks?
We hypothesize that since the dendritic segments are dense and prototype context vectors are sparse (as most pixels in an MNIST image are black), a single segment can learn to identify multiple context vectors, and thus there can be far fewer dendritic segments than unique context vectors.

\end{document}